\documentclass[a4paper,fleqn]{cas-sc}
\usepackage[numbers,sort&compress]{natbib}
\usepackage{booktabs}
\usepackage{graphicx}
\usepackage{amsmath}
\usepackage{array}

\makeatletter
\@for\um@n:={=,<,>,gets,simeq,equal,le,ge,greater,forksnot}\do{%
  \expandafter\let\csname not\um@n\endcsname\@undefined}
\makeatother
\usepackage{unicode-math}
\usepackage{capt-of}

\usepackage[english, bidi=default]{babel}
\babelprovide[import]{arabic}
\babelfont[arabic]{rm}[Script=Arabic,Path=./fonts/,Extension=.ttf,UprightFont=*-Regular,BoldFont=*-Bold,ItalicFont=*-Italic,BoldItalicFont=*-BoldItalic]{Amiri}
\babelfont[arabic]{sf}[Script=Arabic,Path=./fonts/,Extension=.ttf,UprightFont=*-Regular,BoldFont=*-Bold,ItalicFont=*-Italic,BoldItalicFont=*-BoldItalic]{Amiri}
\newcommand{\textarabic}[1]{\foreignlanguage{arabic}{\rmfamily #1}}

\usepackage[section]{placeins}

\begin{document}
\let\WriteBookmarks\relax
\def\floatpagepagefraction{1}
\def\textpagefraction{.001}

\shorttitle{What Majority Labels Hide: Annotator Disagreement and Model Errors in Multi-Dialect Arabic Sexism Detection}
\shortauthors{A. Bazaid and A. Zubiaga}

\title [mode = title]{What Majority Labels Hide: Annotator Disagreement and Model Errors in Multi-Dialect Arabic Sexism Detection}

\author[1]{Alaa Bazaid}[orcid=0009-0007-5275-0734]
\cormark[1]
\ead{a.z.o.bazaid@qmul.ac.uk}
\credit{Conceptualization, Data curation, Methodology, Software,
Formal analysis, Investigation, Visualization, Writing -- original draft,
Writing -- review \& editing}

\author[1]{Arkaitz Zubiaga} [orcid=0000-0003-4583-3623]
\ead{a.zubiaga@qmul.ac.uk}
\credit{Conceptualization, Methodology, Supervision,
Writing -- review \& editing}

\affiliation[1]{organization={School of Electronic Engineering and Computer
Science, Queen Mary University of London},
    addressline={Mile End Road},
    city={London},
    postcode={E1 4NS},
    country={United Kingdom}}

\cortext[1]{Corresponding author}

\begin{abstract}
Judgements of sexism are subjective and may vary across communities, yet sexism detection datasets usually reduce annotator judgements to a single majority label, obscuring disagreement and how models behave on contested cases. We introduce MuDAS, a dataset of YouTube comments in Saudi, Egyptian and Moroccan Arabic, each labelled by three native speakers of its dialect, with the individual annotations preserved. We use it to evaluate Arabic encoders fine-tuned with hard-label, soft-label and perspectivist objectives, GPT-5.6 Terra, and four Arabic-centric large language models. Model errors are more frequent on comments with annotator disagreement. Despite relatively close Macro-F1 scores, GPT-5.6 Terra and the fine-tuned encoders exhibit contrasting error patterns: GPT-5.6 Terra, even with the annotators' definition of sexism, over-classifies gender-related content such as religious advice on modesty, whereas the encoders miss implicit and prescriptive sexism. The Arabic-centric models also over-predict sexism and remain below the encoders. Disagreement-aware learning preserves Macro-F1, but soft labels improve alignment with annotators on contested comments at the expense of unanimous ones, and neither paradigm reliably identifies contested comments. Lower performance on Egyptian Arabic, clearest for the large language models, is not fully explained by the examined measures of test-set composition and dialectness. Majority-label scores can therefore hide contrasting error patterns that
matter for content moderation.
\end{abstract}


\nonumnote{\textbf{Content warning:} This paper contains examples of sexist
and offensive language, quoted from the dataset for the purpose of analysis.
These examples do not reflect the views of the authors.}

\maketitle

\section{Introduction}
\label{sec:intro}

Sexist content on social media platforms is increasingly moderated with the
help of automatic classifiers, which are trained and evaluated on comments
labelled by human annotators \cite{samory2021call,jiang2022swsr}. Whether a comment is sexist, however, is often
a matter of judgement \cite{akhtar2020modeling,yin2023annobert}: annotators may agree that an explicit insult directed
at women is sexist, yet disagree on a religious prescription about women's
dress, a joke, or a statement about the roles of men and women. Such judgements depend in part on who makes them: annotators' beliefs and
identities can shape what they consider offensive \citep{sap2022annotators},
and annotators from different countries or communities label the same content differently,
even when they share a language \citep{lee2024exploring}. Because
interpretations of gender roles and expectations are shaped by social and
cultural norms, what is perceived as sexist may also vary across communities. The standard practice of aggregating annotations into a single
majority label discards this disagreement. A comment labelled sexist by two
of three annotators is then treated in the same way as one labelled sexist by
all three, and a model that agrees with the third annotator is counted as
wrong. Previous work has shown that classifier performance depends strongly
on the level of agreement among annotators \citep{leonardelli2021agreeing},
and has proposed ways of training models that retain this disagreement
\citep{uma2021learning}. 

In the historically prevalent \textit{hard-label} approach, each
comment is assigned its majority label, and disagreement is discarded before
training. With \textit{soft labels}, the model is instead trained on the
distribution of the annotators' labels: a comment labelled sexist by two of
three annotators has a target of two-thirds sexist, so the model learns how
much the annotators disagreed \citep{uma2021learning, fornaciari2021beyond}.
In the \textit{perspectivist} approach, the model predicts the label of each
annotator separately, typically with one output per annotator, so it also
retains which annotator disagreed \citep{davani2022dealing}. The different approaches to modelling a subjective task such as sexism detection raise questions about which approach to choose and the impact of this choice on the result.

These questions have received little attention in the context of the Arabic language, which is spoken across multiple countries with different cultural backgrounds and dialectal variations. Arabic sexism and
misogyny datasets are drawn primarily from Twitter \citep{bensalem2024toxic},
and the only Arabic sexism dataset built from YouTube that we are aware of
covers a single dialect \citep{guellil2021sexism}. ArMIS
\citep{almanea2022armis} preserves individual annotations and shows that
judgements of misogyny differ between annotators, but it contains under a
thousand tweets in mixed dialects, labelled by the same three annotators, hence leaving room for further investigation.
Arabic dialects differ in vocabulary and in the communities that speak them,
and judgements of what is sexist may differ across these communities. No
existing Arabic dataset combines several dialects, annotators who are native speakers of the dialect they label, and
individual annotations. At
the same time, general-purpose and Arabic-centric large language models
(LLMs) are increasingly used for detection without task-specific training,
but their errors on Arabic sexism, and on comments on which annotators
disagree, have not been examined.

In this paper, we introduce MuDAS, a dataset of 6,731 YouTube comments in Saudi,
Egyptian and Moroccan Arabic, each labelled for sexism by three native
speakers of the corresponding dialect, with the individual annotations preserved. We use
it to examine what majority-label evaluation hides, by comparing Arabic encoders fine-tuned under each of these three paradigms
with a general-purpose LLM and four Arabic-centric LLMs prompted under the
same three paradigms. We address
three research questions:

\begin{description}
\item[RQ1] How are the errors of these models distributed across levels of
annotator agreement, and do models differ in the kinds of errors they make
beyond what Macro-F1 captures?
\item[RQ2] Does learning from soft labels or from individual annotators
change how well model predictions align with the annotators, and on which
comments?
\item[RQ3] Does model performance differ across the three dialects and, if
so, to what extent are these differences related to the composition of each
test set (its proportion of sexist and of contested comments) and to the
dialectness of its comments?
\end{description}

Our study makes the following main contributions:

\begin{itemize}
\item We develop and release MuDAS, a dataset of 6,731 YouTube comments in Saudi, Egyptian and
Moroccan Arabic, each annotated by three dialect-matched native speakers,
with individual judgements preserved. By retaining annotator-level labels
across three dialect communities, MuDAS enables the study of disagreement within each community and the comparison of disagreement patterns across
communities. Since each comment is labelled only by annotators of its own
dialect, it also provides a foundation for comparing how different
communities judge the same comments, through additional annotations. We release the dataset and report baselines covering hard-label, soft-label and perspectivist approaches
(Sections~\ref{sec:dataset} and~\ref{sec:res-main}).
\item We conduct comprehensive experiments, among others showing evidence that model errors fall disproportionately on contested
comments; we also observe that GPT-5.6 Terra and the fine-tuned encoders, despite achieving
relatively close Macro-F1 scores, err in opposite directions: GPT-5.6 Terra
over-flags gender-related content such as religious prescriptions on modesty
and neutral debate on gender roles, whereas fine-tuned encoders miss
implicit, prescriptive and benevolent sexism
(Sections~\ref{sec:res-contested} and~\ref{sec:err-direction}--\ref{sec:err-qualitative}).
This over-flagging persists, although at a reduced level, when GPT-5.6 Terra
uses the annotators' definition of sexism against women
(Appendix~\ref{app:definition}).
\item We perform a comparison of the three learning paradigms by agreement level,
showing that, for the fine-tuned encoders, soft-label training largely
preserves Macro-F1 scores but improves alignment with the annotators on contested
comments at a cost on unanimous ones, and that neither soft-label nor
perspectivist models reliably identify which comments are contested
(Sections~\ref{sec:res-main}--\ref{sec:res-contested}).
\item An analysis showing that Arabic-centric LLMs over-predict sexism and
remain below the fine-tuned encoders, and that lower performance on Egyptian
Arabic, clearest for these models, is not fully accounted for by the
composition and the dialectness of the test set
(Sections~\ref{sec:err-direction} and~\ref{sec:err-egyptian}).
\end{itemize}

The remainder of the paper is organised as follows.
In Section~\ref{sec:related}, we review related work on annotator disagreement,
Arabic sexism datasets and LLMs for sexism detection.
Next in Section~\ref{sec:dataset}, we describe the collection and annotation of the
dataset, and follow in Section~\ref{sec:setup} with the models, learning paradigms and
evaluation measures. We then report results in Section~\ref{sec:results}, and analyse the error in more detail in Section~\ref{sec:error-analysis}.
We discuss  the findings, their implications for
content moderation and the limitations of the study in Section~\ref{sec:discussion}, and
we conclude the paper in Section~\ref{sec:conclusion}.

\section{Related Work}
\label{sec:related}

\subsection{Annotator Disagreement}
\label{sec:rw-general}
 
Annotated datasets have traditionally been built on the assumption that each
item has a single correct label, with disagreement among annotators treated
as noise to be removed by aggregation. This assumption has been increasingly
questioned. \citet{aroyo2015truth} argued that disagreement often reflects
genuine ambiguity in the items rather than annotator error, and
\citet{pavlick2019inherent} showed that disagreement in natural language
inference persists as more ratings are collected and cannot be attributed to
noise alone. \citet{plank2022problem} describes such variation as human
label variation, which affects how data are collected, how models are
trained and how they are evaluated. The perspectivist approach goes further
and argues for methods that preserve the divergence of annotators' opinions
rather than reducing it to a single label, not only in the most subjective
tasks \citep{cabitza2023toward}. In subjective tasks, disagreement also
depends on who the annotators are: their beliefs and identities can shape
what they consider offensive \citep{sap2022annotators}.
 
These views have also changed how models are evaluated.
\citet{basile2021we} argued that evaluation should take disagreement into
account rather than rely on a single gold label, and the Learning with
Disagreements shared tasks compare predicted and observed label
distributions with soft metrics such as the Manhattan distance, and evaluate
annotator-level predictions with error rates
\citep{leonardelli2025lewidi, rizzi2024soft}.

\subsection{Learning from Annotator Disagreement}
\label{sec:rw-disagreement}

Methods that learn from disaggregated labels rather than from the majority
label fall into the three families introduced in Section~\ref{sec:intro}
\citep{uma2021learning}: training on the majority label (hard labels), on the
distribution of labels (soft labels) \citep{fornaciari2021beyond}, and on the
labels of individual annotators (perspectivist models), for example with one
output per annotator \citep{davani2022dealing} or with separate models for
groups of annotators whose judgements are similarly polarised
\citep{akhtar2020modeling}. \citet{leonardelli2021agreeing} showed that the
level of agreement among annotators strongly affects classifier performance
in offensive language detection, and that items with low agreement are not
necessarily the result of poor annotation.

A related line of work represents the annotators explicitly. AnnoBERT
\citep{yin2023annobert} learns a representation of each annotator from their
characteristics, combines it with the text of the label the annotator chose,
and at test time predicts the aggregated label; it improved hate speech
detection in particular on the minority class and on items in which
annotators disagreed. \citet{deng2023you} likewise add learned embeddings of
each annotator and of their annotations to the input of a transformer, which
improved performance on six of the eight datasets of their benchmark, and
jury learning \citep{gordon2022jury} models each annotator so that a
prediction can be derived from an explicitly chosen panel of annotators
rather than from the majority. Since each dialect in our data is labelled by
the same three annotators, our perspectivist models follow the simpler
design of one output per annotator \citep{davani2022dealing}.

Previous work has also linked these approaches to ambiguous items:
\citet{fornaciari2021beyond} suggest that the benefit of soft labels is
especially likely for ambiguous instances, and \citet{anand2024don} found that
a multi-annotator model was more confident in minority labels where a
majority-label model was uncertain. To our knowledge, however, the effect of
each paradigm on contested and unanimous items has not been measured
separately, nor compared with the behaviour of LLMs on the same items.

\subsection{Sexism Detection and Disagreement across Languages}
\label{sec:rw-sexism}
 
Sexism detection has been studied in several languages, with datasets for
English \citep{samory2021call, kirk2023semeval}, French
\citep{chiril2020annotated}, Danish \citep{zeinert2021annotating},
Chinese \citep{jiang2022swsr} and Arabic (Section~\ref{sec:rw-arabic}),
among others. These datasets differ in their definitions of sexism and in
their label taxonomies: \citet{samory2021call}, for example, derived their
codebook from psychological scales of sexism, and
\citet{zeinert2021annotating} describe how their taxonomy was refined
through iterative annotation. The EDOS shared task \citep{kirk2023semeval}
introduced a hierarchical taxonomy of sexism in English, which the
definition used in our guidelines builds on (Section~\ref{sec:ds-annotation}).
 
Disagreement in sexism annotation has been addressed most directly in the
EXIST shared tasks on English and Spanish tweets. In EXIST 2023, each tweet
was labelled by six annotators selected to vary in gender and age, the
individual labels were released, and systems were evaluated against both
the majority label and the distribution of labels
\citep{plaza2023overview}. When hard outputs were evaluated against the
distribution of labels, scores were lower than for soft outputs (57.3
against 64.2 for the best system on sexism identification), and the
highest-ranked team under hard evaluation ranked only 14th to 16th under
soft evaluation. The organisers note that whether a tweet is perceived as
sexist may depend on the gender, age and cultural background of the
annotator, but annotators were selected to vary in gender and age only.
Studies of disagreement in labelling data for sexism detection have thus focused mainly on
English and Spanish, while ArMIS remains, to our knowledge, the only Arabic
resource that preserves individual annotations of misogyny
\citep{almanea2022armis}.

\subsection{Arabic Sexism Datasets and Dialects}
\label{sec:rw-arabic}

Most Arabic sexism and misogyny datasets are built from Twitter, including
ArMI \citep{mulki2021armi}, Let-Mi \citep{mulki2021let}, ArMIS
\citep{almanea2022armis} and the dataset of \citet{alhayan2024detection}. In a
review of 54 Arabic toxic-language datasets, \citet{bensalem2024toxic} found
that more than half were drawn from Twitter. The only Arabic sexism dataset
built from YouTube that we are aware of covers a single dialect
\citep{guellil2021sexism}. Table~\ref{tab:arabic-datasets} compares these datasets with MuDAS. 

\begin{table}[htbp]
\centering
\small
\setlength{\tabcolsep}{4pt}
\begin{tabular}{lllrcc}
\toprule
\textbf{Dataset} & \textbf{Source} & \textbf{Dialect(s)} & \textbf{Size} & \textbf{Ann./item} & \textbf{Indiv.} \\
\midrule
Let-Mi \citep{mulki2021let}           & Twitter & Levantine                    & 6,550  & 3 & No \\
ArMI \citep{mulki2021armi}            & Twitter & Gulf, Egyptian, Levantine    & 9,833  & n/r & No \\
\citet{guellil2021sexism}             & YouTube & Algerian                     & 3,798  & 3 & No \\
ArMIS \citep{almanea2022armis}        & Twitter & Mixed                        & 964    & 3 & Yes \\
\citet{alhayan2024detection}          & X       & Not specified (Saudi context) & 7,487 & 1 & No \\
\midrule
\textbf{MuDAS (ours)}                          & YouTube & Saudi, Egyptian, Moroccan    & 6,731  & 3 & Yes \\
\bottomrule
\end{tabular}
\caption{Existing Arabic sexism and misogyny datasets. \textit{Size}: number of
annotated items; \textit{Ann./item}: annotators per item;
\textit{Indiv.}: whether the individual annotations are kept, so that
disagreement can be studied. In MuDAS, each dialect is labelled by three
native speakers of that dialect. \citet{guellil2021sexism} kept only the
comments on which all three annotators agreed, and \citet{alhayan2024detection}
had each tweet labelled by one of six annotators. n/r: not reported.}
\label{tab:arabic-datasets}
\end{table}

ArMIS \citep{almanea2022armis} is the closest resource to ours: it preserves
the labels of three annotators who differ in gender and self-reported degree
of religious belief, and shows that their judgements of misogyny differ
systematically. ArMIS, however, contains 964 tweets in a mix of Arabic
dialects, all labelled by the same three annotators. MuDAS differs in
three respects: it is drawn from YouTube, it is larger (6,731 comments), and
it separates three dialects, each labelled by three native speakers of that
dialect, so that disagreement can be studied within each dialect community.

Arabic varies considerably across dialects \cite{abd1987cross}, and the degree of dialectness of
a text can be estimated with ALDi \citep{keleg2023aldi}. Sexism judgements may
also depend on the cultural background of annotators, yet existing Arabic
datasets rarely match annotators to the dialect of the comments they label.
No existing dataset combines several dialects, dialect-matched annotators
and individual annotations, which is required to study disagreement within
and across Arabic-speaking communities.

\subsection{LLMs for Sexism and Hate Speech Detection}
\label{sec:rw-llms}

LLMs are increasingly used to detect hate speech without task-specific
training: with zero-shot prompting, recent models have performed comparably
to, and sometimes better than, fine-tuned models on several hate speech
benchmarks \citep{plaza2023respectful}, although their performance depends
strongly on the prompt and on the information it includes, such as the
community targeted by a post \citep{roy2023probing}. Evaluations of LLMs on
offensive language with annotation disagreement report competitive performance
on unanimous items but a drop on items with low agreement, with a tendency
to label ambiguous content as offensive \citep{lu-etal-2025-llm}.

Several Arabic-centric LLMs have been released, among them ALLaM
\citep{bari2025allam}, Jais \citep{sengupta2023jais} and its successor Jais~2
\citep{anwar2026jais}, Fanar \citep{team2025fanar} and the Moroccan-adapted
Atlas-Chat \citep{shang2025atlas}, but their behaviour on Arabic sexism, and
on comments on which annotators disagree, has not been evaluated.

\section{Dataset}
\label{sec:dataset}

We introduce MuDAS (\textbf{Mu}lti-\textbf{D}ialect \textbf{A}rabic
\textbf{S}exism), a dataset of 6,731 YouTube comments in Saudi, Egyptian and
Moroccan Arabic, each labelled for sexism by three native speakers of their corresponding
dialect. MuDAS preserves the individual annotations of each annotator instead of aggregating them into a single, unified label. This section describes how the comments were collected
(Section~\ref{sec:ds-collection}), sampled (Section~\ref{sec:ds-sampling}) and
annotated (Section~\ref{sec:ds-annotation}), and summarises the final dataset
(Section~\ref{sec:ds-final}). Figure~\ref{fig:pipeline} summarises the construction of the dataset.

\begin{center}
\includegraphics[width=0.8\textwidth,height=0.45\textheight,keepaspectratio]{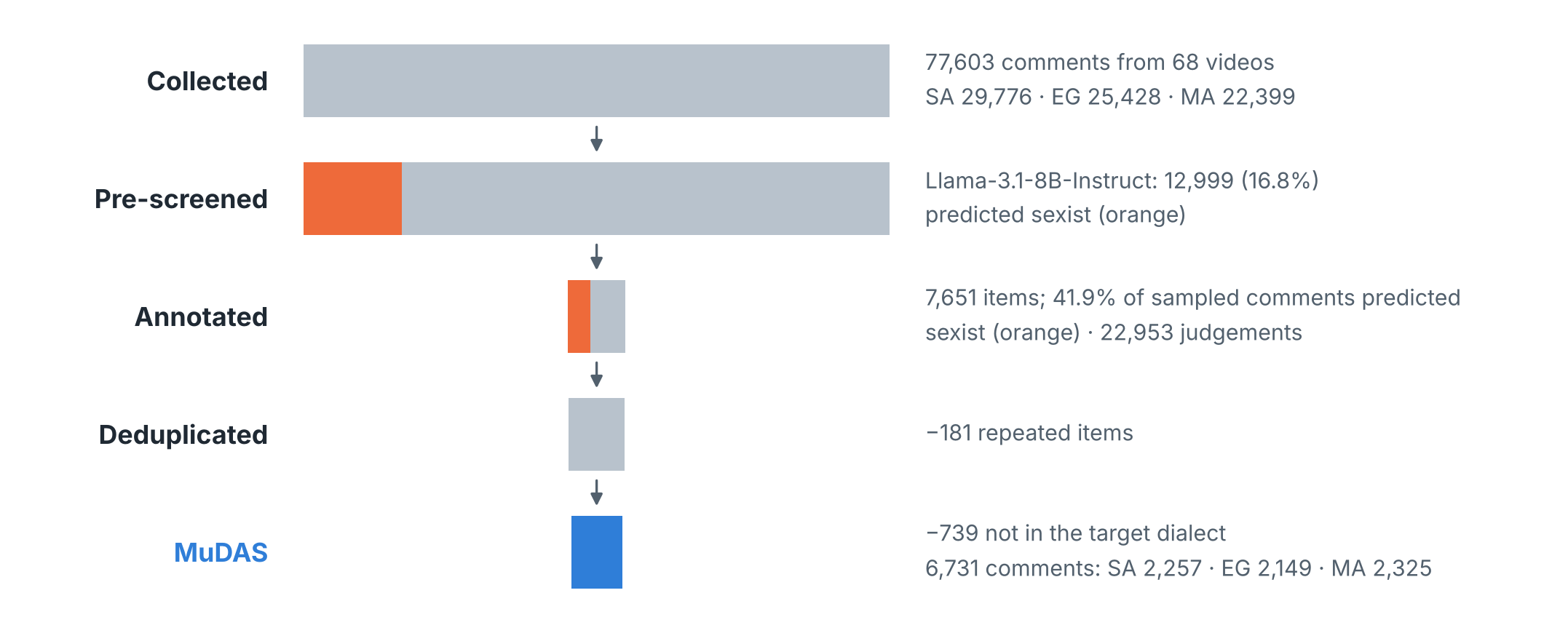}
\captionof{figure}{Construction of MuDAS. Bar widths are proportional to the
number of comments at each stage; the orange segment marks the comments
predicted as sexist by the pre-screening model, which made up 41.9\% of the
sampled comments.}
\label{fig:pipeline}
\end{center}

\subsection{Data collection}
\label{sec:ds-collection}

We collected comments from YouTube, which is widely used in the three target
countries but rarely used as a source of Arabic sexism data
(Section~\ref{sec:rw-arabic}), and whose official Data API
v3\footnote{\url{https://developers.google.com/youtube/data}}
gives free access to public comments. To avoid biasing the data towards explicit abuse, videos
were retrieved by topic rather than by slurs. We compiled 32 search keywords (Appendix~\ref{app:keywords})
on topics that commonly trigger gendered discussion in Arabic-speaking
societies, grouped into five themes: marriage and divorce, work and
independence, appearance and behaviour, roles and relationships, and
controversial issues such as harassment and inheritance. Retrieving by topic
yields sexist and non-sexist comments within the same discussions, so that the
negative class is not trivially off-topic.

For each dialect, videos were selected manually if their subject matched a
keyword and their main speakers used the target dialect. Not every keyword
yielded a suitable video, but all five themes are represented in each
dialect. In total, 68 videos were selected (25 Saudi, 24 Egyptian and 19
Moroccan; Table~\ref{tab:app-videos}). All top-level comments
and replies were collected between June and August 2026, yielding 77,603
comments. 

\subsection{Pre-screening and sampling}
\label{sec:ds-sampling}

Sexist comments make up a small share of YouTube comments, so random sampling
would yield a highly imbalanced annotation set. Keyword filtering, the most
common alternative, captures mainly explicit abuse and biases the data
towards the chosen terms. Following earlier work that used model predictions
to select data for annotation \citep{kirk2023semeval, leonardelli2021agreeing},
we pre-screened all comments with Llama-3.1-8B-Instruct, prompted in Arabic
with a short definition of sexism that covers direct and implicit forms and
asked to answer with a single label (Appendix~\ref{app:prescreen}). Of the
77,603 comments, 16.8\% were predicted as sexist (18.6\% for Saudi, 15.0\%
for Egyptian and 16.3\% for Moroccan).
These predictions were used only to guide sampling and were never shown to
annotators.

Each dialect was annotated in 11 batches of about 240 comments each (the first batch included only 120 comments as a trial). Comments predicted as unclear were grouped with those predicted as not
sexist, and comments predicted as sexist were sampled at a higher rate than
their share of the collected data. Because annotators judged most comments in
the early batches as not sexist, this rate was increased in later batches. Of
the sampled comments, excluding gold and repeated items, 41.9\% had been
predicted as sexist, against 16.8\% of all collected comments.

Retaining a substantial share of comments not predicted as sexist
limits the risk that the dataset contains only what the pre-screening model
already recognises as sexist. Because users from different countries comment on the same videos, each comment was assigned to a dialect based on its own language rather than that
of the video, and annotators confirmed whether each comment was written in
their dialect (Section~\ref{sec:ds-final}).

\subsection{Annotation}
\label{sec:ds-annotation}

\paragraph{Annotators.}
Annotators were recruited through Prolific\footnote{\url{https://www.prolific.com}}
and completed all annotation tasks on
Qualtrics,\footnote{\url{https://www.qualtrics.com}} accessed through Prolific.
Since the dataset is designed to capture how sexism is judged within each
dialect community, annotators had to share the linguistic and cultural
background of the comments they labelled. Prolific's screening filters were
therefore used to require three criteria, each matching the target country
(Saudi Arabia, Egypt or Morocco): nationality, country of birth and current
country of residence. Requiring all three excludes participants who live in
the country without being raised there, as well as nationals born or living
abroad. These criteria were introduced after a first pilot round in which
eligibility was based on residence alone, which admitted participants whose
annotations diverged markedly from those of the others, including a
participant of a different nationality and a national born and living outside
the region.

After three pilot rounds, three annotators were selected per dialect, with
the constraint that each group included at least one woman, given evidence
that annotator gender can influence judgements of sexist content
\citep{almanea2022armis}. Each of the final groups comprised two men and one
woman, aged 20 to 31. The same nine annotators labelled all comments, with no
withdrawals or replacements, so that every comment within a dialect was
labelled by the same three native speakers of that dialect.

\paragraph{Guidelines.}
Separate guidelines were written for each dialect, sharing the same structure
but using examples from the dialect in question. Sexism was defined as any expression that
demeans women, discriminates against them, reinforces stereotypes about them
or justifies harm done to them, whether directed at women as a group or at an
individual woman because she is a woman. The definition builds on that of
EDOS \citep{kirk2023semeval} and extends it to benevolent sexism
\citep{glick1996ambivalent} and to the denial of discrimination against women.

The guidelines describe eight types of sexism: stereotyping and
objectification, insult and belittling, dominance, violence and harassment,
endorsement and justification, doubt and victim blaming, benevolent sexism,
and ideological discourse and inequality. They also describe five categories
of comments that mention women without being sexist: praise of women,
neutral text, criticism of a specific woman unrelated to her gender, a
woman's own account of mistreatment, and criticism of discrimination. Each
type is illustrated with the same example in the three dialects
(Appendix~\ref{app:guideline-examples}), and minimal pairs contrast similar
sexist and non-sexist comments. The types served to illustrate the
definition: annotators assigned a binary label and did not label the type. For each comment, annotators also indicated whether it was written in their
own dialect, so that comments in another dialect could be eventually excluded from the sample
(Section~\ref{sec:ds-final}).

\paragraph{Quality control.}
Each batch of about 240 comments was divided into four annotation sessions of about 60 comments (two sessions in the first batch of 120). Two kinds of
check were included in the sessions. First, each session contained hidden
gold items \citep{oleson2011programmatic}: real comments whose label follows
clearly from the guidelines (450 in total, balanced between the two classes),
used as ongoing qualifying tests. Individual accuracy on these items ranged
from 95.3\% to 100\% (mean 98.3\%), and the majority label matched the gold
label on all 450. Second, from the fourth batch, each session repeated one
comment from an earlier batch to measure self-consistency; repeating items
from earlier batches rather than within the same session can reduce the
effect of memory on the second judgement \citep{abercrombie2023temporal}.
Individual annotators agreed with their earlier label on 93.4\% of the 96
repeated comments on average (mean Cohen's $\kappa = 0.86$), and the
majority label was unchanged for 95.8\% of them. Since all annotators
performed well on both checks, none was excluded.

\subsection{Final dataset}
\label{sec:ds-final}

In total, 7,651 items were annotated. Repeated items were reduced to their
first occurrence (181 items: the 96
self-consistency repeats and 85 unplanned repeats), and comments that fewer than two of the three
annotators confirmed as written in the target dialect were excluded (739
comments). The final dataset contains 6,731 comments: 2,257 Saudi, 2,149
Egyptian and 2,325 Moroccan. The 414 gold items that passed the dialect filter
were retained, since they are genuine comments annotated under the same
conditions, and are flagged in the data.
Table~\ref{tab:ds-stats} summarises the final dataset.

\begin{table}[htbp]
\centering
\small
\begin{tabular}{lcccc}
\toprule
 & \textbf{Saudi} & \textbf{Egyptian} & \textbf{Moroccan} & \textbf{All} \\
\midrule
Videos                          & 25    & 24    & 19    & 68 \\
Comments (final)                & 2,257 & 2,149 & 2,325 & 6,731 \\
Sexist (majority label)         & 42.7\% & 34.5\% & 49.2\% & 42.3\% \\
\midrule
\multicolumn{5}{l}{\textit{Annotators labelling the comment sexist ($s$)}} \\
\quad 0/3                       & 37.8\% & 50.4\% & 37.1\% & 41.6\% \\
\quad 1/3                       & 19.5\% & 15.1\% & 13.6\% & 16.1\% \\
\quad 2/3                       & 18.3\% & 14.5\% & 21.2\% & 18.1\% \\
\quad 3/3                       & 24.4\% & 20.1\% & 28.0\% & 24.3\% \\
Unanimous                       & 62.2\% & 70.5\% & 65.2\% & 65.9\% \\
\midrule
Fleiss' $\kappa$                & 0.486 & 0.565 & 0.533 & 0.532 \\
Gwet's AC1                      & 0.506 & 0.640 & 0.537 & 0.557 \\
Annotator sexist rate (range)   & 36.7--49.2\% & 32.4--36.2\% & 33.2--55.3\% & 32.4--55.3\% \\
\bottomrule
\end{tabular}
\caption{Statistics of the final dataset. $s$ is the proportion of the three
annotators who labelled a comment sexist; unanimous comments are those with
$s = 0$ or $s = 1$, and contested comments those with
$s \in \{\tfrac{1}{3}, \tfrac{2}{3}\}$. Agreement is computed on the sexism label.}
\label{tab:ds-stats}
\end{table}

Agreement among the three annotators was moderate, with Fleiss' $\kappa$ of
0.532 and Gwet's AC1 of 0.557, close to the $\kappa$ of 0.525 reported for
ArMIS \citep{almanea2022armis}. The annotators agreed fully on 65.9\% of the
comments, which we call \textit{unanimous} ($s = 0$ or $s = 1$), and split on
the remaining 34.1\% (Figure~\ref{fig:dataset}a). We call the latter
\textit{contested} comments: those labelled sexist by one or two of the three
annotators ($s \in \{\tfrac{1}{3}, \tfrac{2}{3}\}$). The dialects differed in both
respects: the Egyptian comments had the lowest proportion of sexist comments
(34.5\%) and the highest proportion of unanimous ones (70.5\%), while the
Moroccan comments were the most balanced (49.2\% sexist). 

\begin{center}
\includegraphics[width=0.85\textwidth,height=0.45\textheight,keepaspectratio]{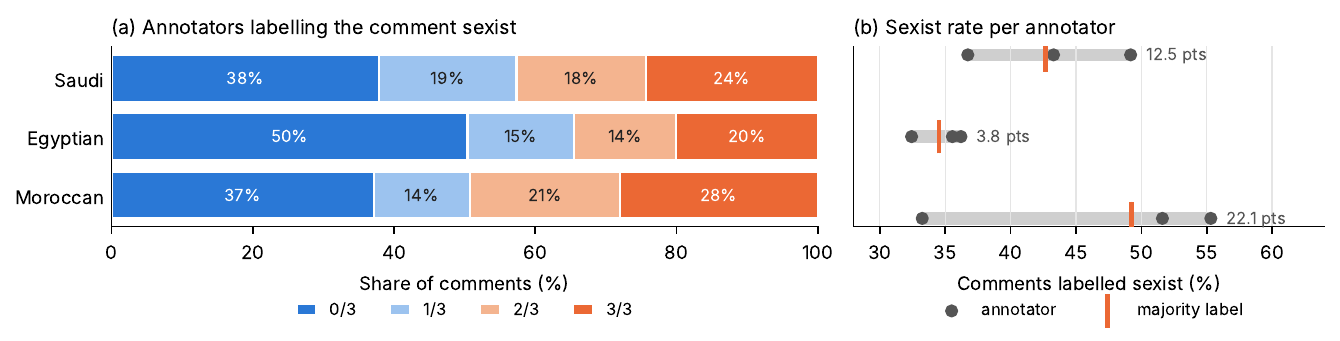}
\captionof{figure}{Annotator agreement in each dialect. (a) Distribution of
the number of annotators, out of three, who labelled a comment sexist.
(b) Proportion of comments labelled sexist by each annotator (dots) and by the
majority label (vertical line); the grey bar spans the three annotators of
each dialect.}
\label{fig:dataset}
\end{center}

Annotators also
differed in how often they assigned the sexist label, from 32.4\% to 55.3\%;
this spread was narrowest within the Egyptian group (3.8 percentage points)
and widest within the Moroccan group (22.1 points) (Figure~\ref{fig:dataset}b). Since all annotators were accurate on the gold items, these differences are
unlikely to reflect inattentive annotation, and may instead reflect
different thresholds for judging a comment sexist, although the gold items
are clear cases and do not test the boundary.

\section{Experimental Setup}
\label{sec:setup}

\subsection{Data splits and training settings}
\label{sec:setup-splits}

Each dialect was split once into a training pool (80\%) and a test set (20\%),
stratified by the majority label. For the encoders, 15\% of each training pool
was further held out for checkpoint selection. The same test sets were used in
every experiment (Table~\ref{tab:splits}). Two training settings were compared.
In the \textit{in-dialect} setting, a model was trained and tested within a
single dialect. In the \textit{mixed-dialect} setting, the three training pools
were combined, and the model was tested on the combined test set of 1,347
comments.

The test sets include 90 gold items (6.7\% of the combined test set; 7.3\%,
5.1\% and 7.5\% of the Saudi, Egyptian and Moroccan test sets), of which 46 are sexist and 44 not sexist. These are real comments used for quality control during annotation; like all other comments, they are evaluated against the labels of the three annotators, whose majority matched the gold label on all of them. Since gold items were
chosen as clear cases, Appendix~\ref{app:nogold} repeats the main analyses
without them.

\begin{table}[htbp]
\centering
\small
\begin{tabular}{lcccc}
\toprule
 & \textbf{Saudi} & \textbf{Egyptian} & \textbf{Moroccan} & \textbf{Mixed} \\
\midrule
Training pool               & 1,805 & 1,719 & 1,860 & 5,384 \\
\quad Training (encoders)   & 1,534 & 1,461 & 1,581 & 4,576 \\
\quad Validation (encoders) &   271 &   258 &   279 &   808 \\
Test                        &   452 &   430 &   465 & 1,347 \\
\bottomrule

\end{tabular}
\caption{Number of comments in each split. The mixed-dialect split is the
union of the three dialect splits.}
\label{tab:splits}
\end{table}

\subsection{Preprocessing}
Comments were minimally preprocessed, using the same procedure for the text shown to annotators, the model inputs and the data. Text was
normalised to Unicode NFC, invisible and control characters were removed,
whitespace and line breaks were collapsed, and URLs and email
addresses were replaced with the placeholders \texttt{URL} and
\texttt{EMAIL}. To support anonymised processing of the data, all user mentions were replaced with \texttt{@USER}. No stemming, stop-word removal or orthographic normalisation of
Arabic letters was applied, and diacritics, elongation, emojis and hashtags
were kept, since they may signal dialect, emphasis or sarcasm.

\subsection{Learning paradigms}
\label{sec:setup-paradigms}

We compared three paradigms that retain increasing amounts of the annotators'
disagreement \citep{uma2021learning}. All three use the same encoder architecture and data splits, but differ in
their training targets and, for the perspectivist paradigm, in the output
layer.

\begin{itemize}
  \item \textbf{Hard labels.} Each comment is assigned its majority label (i.e. a unified sexist or not sexist for the label selected by at least two of the annotators), and
  the model is trained with cross-entropy. Disagreement is resolved before
  training.
  \item \textbf{Soft labels.} The target is the distribution $[1-s, s]$, where
  $s$ is the proportion of annotators who labelled the comment sexist (0, 1/3, 2/3, or 3/3), and the
  model is trained with soft cross-entropy. This retains how much the
  annotators disagreed.
  \item \textbf{Perspectivist.} A shared encoder is combined with one
classification head per annotator, each trained with cross-entropy on the
labels of its annotator \citep{davani2022dealing}. The model therefore
predicts the label that each individual annotator would assign: for each
test comment, it produces one prediction per annotator of that comment's
dialect, that is, whether that annotator would label the comment sexist.
This retains which annotator disagreed. The three annotator-level
predictions are used directly to evaluate the model against each
annotator's own labels, and are combined into a predicted distribution and
a majority decision for comparison with the other paradigms
(Section~\ref{sec:setup-eval}). In the mixed-dialect setting, the model has
one head for each of the nine annotators, and each comment updates, and is
predicted by, only the heads of its three annotators.
\end{itemize}

No class weighting was applied, so that the paradigms do not also differ in
weighting, and because weighting would shift the predicted probabilities on
which the soft metrics are computed. Each paradigm was selected on its own objective:
validation Macro-F1 for hard labels, validation Manhattan distance (MD) for soft
labels, and the mean validation Macro-F1 of the heads for the perspectivist
models.

\subsection{Models}
\label{sec:setup-models}

\paragraph{Lexical baseline.}
Logistic regression on TF-IDF features, trained with hard labels, serves as a
lexical reference for classification and for Disagreement AUROC.

\paragraph{Encoders.}
We fine-tuned two Arabic encoders in all three paradigms: MARBERT
\citep{abdul2021arbert}, pretrained on dialectal Arabic tweets, and the
Twitter variant of AraBERT \citep{antoun2020arabert}. Both used a maximum
sequence length of 256 tokens, which covers 96--99\% of the comments, a batch
size of 16, AdamW with a weight decay of 0.01, a linear schedule with 10\%
warm-up, a learning rate of $2 \times 10^{-5}$ and up to five epochs. Each
experiment was repeated with three random seeds, and scores are reported as
means over seeds. Where a single decision per comment is needed, as in the error analysis, the
seeds' predicted probabilities are averaged and the decision is obtained by
thresholding this mean at 0.5. This threshold is the same for all models
(Section~\ref{sec:setup-eval}) and was not tuned, so that the paradigms are
compared under a common decision rule.

\paragraph{Large language models.}
We prompted GPT-5.6 Terra \citep{openai2026gpt56}, a proprietary multilingual
model, and four open-weight Arabic-centric models: ALLaM-7B
\citep{bari2025allam}, Fanar-9B \citep{team2025fanar}, Jais-2-8B
\citep{anwar2026jais} and Atlas-Chat-9B \citep{shang2025atlas}, which was
adapted to Moroccan Arabic. The open-weight
models were run once with greedy decoding. GPT-5.6 Terra supports only a
sampling temperature of 1, so each condition was run three times and the
scores averaged.

Each paradigm has a corresponding prompt. The hard-label prompt asks whether
the comment is sexist; the soft-label prompt states that three
Arabic-speaking annotators judged the comment independently and asks how many
of them answered Yes, which gives
$\hat{s} \in \{0, \tfrac{1}{3}, \tfrac{2}{3}, 1\}$; and the perspectivist
prompt asks the model to judge as one specific annotator would, given
examples of that annotator's labels. 

All prompts include the definition of sexism given to the annotators
(Section~\ref{sec:ds-annotation}): a comment is sexist if it demeans women,
discriminates against them, reinforces stereotypes about them, or justifies
harm done to them, whether directed at women as a group or at an individual
woman because she is a woman. The prompts were written in English and then
translated into Arabic; the Arabic prompts use the wording of the Arabic
guidelines. The hard-label and soft-label prompts were run in two versions:
without examples (zero-shot) and with eight examples from the training pool
of the test dialect, or eight from each dialect (24 in total) in the
mixed-dialect setting (few-shot). The perspectivist prompt was run with
examples only, since it relies on examples of each annotator's labels, and
only on the in-dialect test sets, since each prompt refers to the
annotators of a single dialect. The full prompts are given in
Appendix~\ref{app:prompts}.
 
The few-shot examples were drawn from the training pool only, so no test
comment appeared as an example, and were chosen for each paradigm: for the
hard-label prompt, the proportion of sexist examples matched that of the training pool,
and half of each class were contested comments; for the soft-label prompt,
two examples were drawn for each number of sexist votes; and for the
perspectivist prompt, eight examples were drawn per annotator, matched to that
annotator's own rate. The open-weight models generated up to 256 new tokens.
A reply that did not contain a valid answer, including a vote count outside
0--3, was counted as unparseable and excluded from that model's scores.
Unless stated otherwise, LLM results use the English few-shot prompt.

\subsection{Evaluation}
\label{sec:setup-eval}

Evaluation was organised around three questions, each answered by a primary
metric. For each test comment, $s$ is the gold proportion of sexist votes,
$\hat{s}$ the predicted one, and $\hat{h}$ the predicted label. For the
hard-label and soft-label encoders, $\hat{s}$ is the predicted probability of
the sexist class; for the perspectivist models, it is the proportion of the
three annotator-specific predictions that are sexist; for the soft-label
prompts, it is the predicted number of sexist votes divided by three; and for
the hard-label prompts, $\hat{s} = \hat{h}$. In all cases,
$\hat{h} = \mathbb{1}[\hat{s} \geq 0.5]$.

\begin{itemize}
  \item \textbf{Is the decision correct?} \textit{Macro-F1} of $\hat{h}$
  against the majority label.
  \item \textbf{Is the amount of disagreement correct?} The \textit{Manhattan
  distance (MD)} between the gold and predicted distributions,
  $\mathrm{MD} = 2\,|s - \hat{s}|$, averaged over comments. MD ranges from 0
  to 2 and is the soft metric of the third Learning with Disagreements shared
  task for categorical data \citep{leonardelli2025lewidi, rizzi2024soft}.
  \item \textbf{Is disagreement predicted on the right comments?}
  \textit{Disagreement F1} treats contested comments ($s \in \{\tfrac{1}{3},
  \tfrac{2}{3}\}$) as the positive class, and a comment is predicted as
  contested if the predicted number of sexist votes, $\mathrm{round}(3\hat{s})$,
  is 1 or 2. Since marking every comment as contested yields a high recall, it
  is reported with the predicted contested share $\hat{p}$. As a
  threshold-free check, we follow \citet{yelkanat2026decomposing}, who
  evaluated the prediction of contested items with ROC-AUC, and compute
  \textit{Disagreement AUROC}, using $\hat{s}(1-\hat{s})$ as the score, which
  is highest when $\hat{s}$ is close to 0.5. AUROC can be computed from any
score, but when $\hat{s}$ takes only a few values, many comments receive the
same score and the ranking it evaluates consists largely of ties. It is
therefore computed only where $\hat{s}$ takes enough distinct values to rank
comments: for the encoders, using the mean of the heads' predicted
probabilities for the perspectivist models, and for GPT-5.6 Terra with the
soft-label and perspectivist prompts, whose predictions are averaged over
three runs.
\end{itemize}

Cross-entropy is not used as a primary metric: for the LLMs, $\hat{s}$ takes
only the values $0, \tfrac{1}{3}, \tfrac{2}{3}$ and $1$, so a prediction of 0
or 1 on a contested comment makes cross-entropy depend on the constant used to
clip it rather than on the model. We prefer MD to the Jensen--Shannon
divergence because it is directly interpretable as twice the difference in the
proportion of sexist votes, and because JSD fails some of the properties
expected of a soft metric \citep{rizzi2024soft}.

For the perspectivist models, we also report the annotator error rate, the
proportion of incorrect predictions over all comment--annotator pairs
\citep{leonardelli2025lewidi}. Each metric is read against a reference: the
majority class for Macro-F1, a constant prediction equal to the median of $s$
in the training pool for MD, a predictor that marks every comment as
contested for Disagreement F1, and 0.5 for Disagreement AUROC, with the TF-IDF
baseline as a lexical reference. A hard-label prediction cannot express
disagreement, so its MD on a contested comment is at least $\tfrac{2}{3}$.

To estimate uncertainty due to the sample of test comments, 95\% confidence
intervals were computed using 2,000 bootstrap resamples of comments,
stratified by dialect. Model comparisons were paired, using the same
resamples for the models being compared, with confidence intervals computed
directly on the differences. Scores in the results tables are means over
seeds (encoders) or runs (GPT-5.6 Terra). When a confidence interval refers
to these scores, as for the comparison of GPT-5.6 Terra with AraBERT in
Section~\ref{sec:res-main}, the metric was computed for each seed and each
run and then averaged. For the other confidence intervals and the error
analysis, encoder predictions were aggregated by averaging the predicted
probabilities across the three seeds, thresholded at 0.5 where a decision
is needed. For GPT-5.6 Terra, the analyses by agreement level and dialect
(Sections~\ref{sec:res-contested}, \ref{sec:err-direction}
and~\ref{sec:err-egyptian}) computed each metric separately for the three
runs and then averaged the results. During bootstrap resampling, comments
were sampled as units, retaining all three run predictions for each sampled
comment. Majority voting across runs was used where a single decision per
comment was compared with the encoders (the second comparison in
Section~\ref{sec:res-main}, Section~\ref{sec:err-qualitative} and
Appendices~\ref{app:definition}--\ref{app:annotators}). For the soft-label
prompt, predicted vote counts were averaged across runs. Since the encoders
and GPT-5.6 Terra aggregate their predictions differently, comparisons of
these decisions concern aggregated predictions rather than single runs, and
their differences can therefore differ slightly from those between table
means. The intervals reflect the sampling of test comments only, not the
variation between seeds or runs.

\section{Results}
\label{sec:results}

This section and the next address the research questions of
Section~\ref{sec:intro}. We first report how well the models predict the
majority label and the distribution of annotator judgements
(Section~\ref{sec:res-main}), then whether they identify which comments are
contested (Section~\ref{sec:res-disagreement}), and finally how their errors
and their alignment with the annotators differ between contested and
unanimous comments (Section~\ref{sec:res-contested}). These sections address
RQ2, and Section~\ref{sec:res-contested} also the first part of RQ1. The
error analysis (Section~\ref{sec:error-analysis}) completes RQ1 and addresses
RQ3, and the answers to the three questions are discussed in
Section~\ref{sec:discussion}.

\subsection{Classification and distribution alignment}
\label{sec:res-main}

This first set of results compares the models on the two standard ways of
evaluating against annotators' judgements. Macro-F1 measures how well a
model predicts the majority label, as in conventional evaluation, while MD
measures how close its predicted proportion of sexist votes is to the
proportion observed among the three annotators, and therefore also rewards
a model for reflecting disagreement (Section~\ref{sec:setup-eval}). 
For the perspectivist models, which predict the label of each annotator
with one head per annotator on a shared encoder, the three annotator-level
predictions are combined for this comparison: their majority is compared
with the majority label for Macro-F1, and the share of heads predicting sexist is used as $\hat{s}$ for MD. The annotator error rate, which compares each annotator-level prediction
with that annotator's own label, is reported for the LLMs at the end of
this section.

Comparing
the hard-label, soft-label and perspectivist versions of each model shows
whether learning from disagreement changes either measure (RQ2), and
comparing the fine-tuned encoders with the prompted LLMs shows how the two
families differ under the same three paradigms.
Table~\ref{tab:res-main} reports both measures for all models on each
in-dialect test set and on the mixed-dialect test set, together with
simple reference points: a majority-class predictor for Macro-F1, a
constant prediction of the median for MD, and a TF-IDF baseline. 

\begin{table*}[htbp]
\centering
\small
\setlength{\tabcolsep}{3.5pt}
\begin{tabular}{llcccccccc}
\toprule
 & & \multicolumn{4}{c}{\textbf{Macro-F1} $\uparrow$} & \multicolumn{4}{c}{\textbf{MD} $\downarrow$} \\
\cmidrule(lr){3-6}\cmidrule(lr){7-10}
\textbf{Model} & \textbf{Paradigm} & SA & EG & MA & Mix & SA & EG & MA & Mix \\
\midrule
\textit{Majority class} & -- & 0.364 & 0.396 & 0.337 & 0.366 & -- & -- & -- & -- \\
\textit{Constant median} & -- & -- & -- & -- & -- & 0.727 & 0.730 & 0.776 & 0.745 \\
TF-IDF + LR & Hard & 0.686 & 0.563 & 0.728 & 0.702 & -- & -- & -- & -- \\
\midrule
\midrule
\multicolumn{10}{c}{\textbf{Encoders}} \\
\midrule
MARBERT & Hard & 0.750 & 0.745 & 0.743 & 0.759 & 0.518 & 0.515 & 0.566 & 0.516 \\
 & Soft & 0.756 & 0.743 & 0.767 & 0.773 & 0.498 & 0.460 & 0.495 & 0.471 \\
 & Persp. & 0.755 & 0.753 & 0.748 & 0.770 & 0.490 & \textbf{0.436} & 0.498 & 0.449 \\
\addlinespace
AraBERT & Hard & 0.758 & 0.732 & 0.776 & 0.771 & 0.483 & 0.507 & 0.459 & 0.476 \\
 & Soft & 0.756 & 0.716 & 0.767 & 0.776 & 0.477 & 0.523 & 0.464 & 0.453 \\
 & Persp. & 0.761 & 0.733 & 0.778 & 0.777 & 0.482 & 0.460 & 0.454 & 0.445 \\
\midrule
Mean & Hard   & 0.754 & 0.738 & 0.760 & 0.765 & 0.500 & 0.511 & 0.512 & 0.496 \\
 & Soft   & 0.756 & 0.730 & 0.767 & 0.774 & 0.488 & 0.492 & 0.480 & 0.462 \\
 & Persp. & 0.758 & 0.743 & 0.763 & 0.774 & 0.486 & 0.448 & 0.476 & 0.447 \\
\midrule
\midrule
\multicolumn{10}{c}{\textbf{Large Language Models (LLMs)}} \\
\midrule
GPT-5.6 Terra & Hard   & 0.782 & 0.737 & 0.837 & 0.799 & 0.500 & 0.568 & 0.412 & 0.467 \\
              & Soft   & 0.798 & 0.761 & \textbf{0.851} & \textbf{0.809} & \textbf{0.399} & 0.487 & \textbf{0.359} & \textbf{0.398} \\
              & Persp. & \textbf{0.804} & \textbf{0.778} & 0.812 & n/a & 0.410 & 0.467 & 0.406 & n/a \\

\addlinespace
ALLaM         & Hard   & 0.688 & 0.605 & 0.712 & 0.692 & 0.649 & 0.818 & 0.613 & 0.653 \\
              & Soft   & 0.625 & 0.604 & 0.705 & 0.547 & 0.673 & 0.857 & 0.602 & 0.805 \\
              & Persp. & 0.666 & 0.576 & 0.690 & n/a & 0.616 & 0.819 & 0.576 & n/a \\
\addlinespace
Fanar         & Hard   & 0.670 & 0.506 & 0.675 & 0.619 & 0.684 & 1.004 & 0.686 & 0.780 \\
              & Soft   & 0.561 & 0.339 & 0.579$^\dagger$ & 0.647$^\ddagger$ & 0.659 & 1.180 & 0.746$^\dagger$ & 0.723$^\ddagger$ \\
              & Persp. & 0.641 & 0.518 & 0.598 & n/a & 0.683 & 0.983 & 0.761 & n/a \\
\addlinespace
Jais-2        & Hard   & 0.677 & 0.558 & 0.622 & 0.616 & 0.635 & 0.916 & 0.778 & 0.787 \\
              & Soft   & 0.398$^\dagger$ & 0.280$^\dagger$ & 0.375$^\dagger$ & 0.313 & 0.944$^\dagger$ & 1.234$^\dagger$ & 1.005$^\dagger$ & 1.153 \\
              & Persp. & 0.597 & 0.542 & 0.565 & n/a & 0.757 & 0.875 & 0.829 & n/a \\
\addlinespace
Atlas-Chat    & Hard   & 0.676 & 0.577 & 0.742 & 0.671 & 0.685 & 0.870 & 0.576 & 0.695 \\
              & Soft   & 0.633 & 0.637 & 0.692 & 0.653 & 0.661 & 0.720 & 0.597 & 0.667 \\
              & Persp. & 0.706 & 0.628 & 0.688 & n/a & 0.561 & 0.629 & 0.574 & n/a \\

\midrule
Mean & Hard   & 0.699 & 0.597 & 0.718 & 0.679 & 0.631 & 0.835 & 0.613 & 0.676 \\
 & Soft   & 0.603 & 0.524 & 0.640 & 0.580$^{\S}$ & 0.667 & 0.896 & 0.662 & 0.756$^{\S}$ \\
 & Persp. & 0.683 & 0.608 & 0.671 & n/a & 0.605 & 0.755 & 0.629 & n/a \\
 
\bottomrule
\end{tabular}
\caption{Classification (Macro-F1) and distribution alignment (MD) on each
in-dialect test set (SA, EG, MA) and on the mixed-dialect test set (Mix).
The encoders in the Mix column were trained on the pooled data. Encoder
scores are means over three seeds; GPT-5.6 Terra scores are means over three
runs. LLMs use the English few-shot prompt. The best score in each column is
in bold. $^\dagger$~Between 98.7\% and 99.6\% valid replies.
$^\ddagger$~Computed only on the 389 of the 1,347 comments (29\%) with a valid
reply, which include fewer contested comments than the full test set (26\%
against 31\%), so this score is not comparable with the rest of the column; on
the same 389 comments, the hard-label prompt obtained a Macro-F1 of 0.691.
n/a: the perspectivist prompts were not run on the mixed set. The mean rows average the models above; they include the scores marked
$^\dagger$. $^{\S}$~Mean over four LLMs, excluding Fanar ($^\ddagger$).}
\label{tab:res-main}
\end{table*}

\paragraph{Classification.} All models exceeded the majority-class reference.
GPT-5.6 Terra obtained the highest Macro-F1 on every test set with at least one
of its three prompts. With the hard-label prompt, its margin over the
hard-label AraBERT was 0.061 on the Moroccan test set, 0.024 on the Saudi and
0.005 on the Egyptian test set, and 0.028 on the mixed-dialect test set
(0.799 against 0.771). 

On the means over runs and seeds reported in Table~\ref{tab:res-main}, the
95\% confidence interval of this difference excluded zero on the Moroccan
(0.061 [0.020, 0.101]) and the mixed-dialect test set (0.028 [0.003, 0.053]),
but not on the Saudi or Egyptian test set. On aggregated decisions
(Section~\ref{sec:setup-eval}), with one decision per comment obtained by
majority vote over the three runs for GPT-5.6 Terra and by thresholding the
mean of the seeds' probabilities for AraBERT, the difference on the
mixed-dialect test set was 0.023 [$-$0.003, 0.052]. The advantage of
GPT-5.6 Terra on this test set is therefore small and depends on how
predictions are aggregated.

The hard-label Arabic encoders exceeded the TF-IDF
baseline on every test set, by 0.015--0.182, with the largest margin on the
Egyptian test set, where the baseline obtained 0.563. The four Arabic LLMs remained below the Arabic encoders: across all prompt
conditions (Appendix~\ref{app:prompt-effects}), the best score of any Arabic
LLM was below that of MARBERT and AraBERT in every paradigm and on every
test set, by 0.001 on the Moroccan test set (Atlas-Chat, 0.742, against the
hard-label MARBERT, 0.743) to 0.066 on the mixed-dialect test set. On the
mixed-dialect test set, it was also below the TF-IDF baseline (0.693 against
0.702). This comparison reflects the whole set-up rather than the models
alone: the encoders were fine-tuned on about 1,500 comments per dialect,
whereas the LLMs were prompted with 8 examples, or 24 in the mixed-dialect
setting.

\paragraph{In-dialect and mixed-dialect training.}
The mixed-dialect test set consists of the three in-dialect test sets, so
the two training settings can be compared on the same 1,347 comments by
pooling the predictions of the three in-dialect models. Training a single
model on the pooled data gave a slightly higher Macro-F1 than the three
in-dialect models combined, for MARBERT (0.759 against 0.750), AraBERT (0.771
against 0.761) and the TF-IDF baseline (0.702 against 0.688). These
differences are of the same order as the variation across seeds (standard
deviation up to 0.018), so combining the dialects neither clearly helped nor
harmed the encoders, although each dialect accounted for only about a third of the pooled
training data.

\paragraph{Effect of the learning paradigm.}
For each encoder and test set, the three paradigms were within 0.024
Macro-F1 of each other. On the seed-averaged predictions, the 95\% confidence interval of the
difference in Macro-F1 between a disagreement-aware paradigm and the
hard-label paradigm included zero in all 16 comparisons
(Table~\ref{tab:app-ci-paradigm}). Relative to the hard-label encoders, the
perspectivist paradigm reduced MD in all eight encoder--test set combinations
and the soft-label paradigm in six. On the seed-averaged predictions, however,
the interval of the overall difference in MD excluded zero in only five of the
16 comparisons, four of them on the Egyptian test set; in one of these, the
soft-label AraBERT on the Egyptian test set, MD increased and the lower bound
of the interval was 0.0001. 

Averaged over the two encoders and the three in-dialect test sets (mean
rows of Table~\ref{tab:res-main}), the
three paradigms obtained almost the same Macro-F1 (0.751--0.755), while MD
decreased from the hard-label (0.508) to the soft-label (0.486) and the
perspectivist paradigm (0.470). Averaged over the five LLMs, the hard-label
prompt obtained the highest Macro-F1 (0.671, against 0.654 for the
perspectivist and 0.589 for the soft-label prompt), and the perspectivist
prompt the lowest MD (0.663, against 0.693 and 0.742). The perspectivist
paradigm, whose output is closest to the distribution of annotator labels,
therefore gave the lowest MD in both families. The low soft-label average
of the LLMs comes from the Arabic LLMs: for GPT-5.6 Terra alone, the
soft-label prompt was the best on both measures.

Overall, changing how the annotators' judgements are represented, in training or in the prompt, tended to lower MD without a clear change in the quality of the majority decision, although the reduction in MD was itself clear in only a minority of comparisons.
A lower MD, however, does not show that a model
identifies which comments are contested. MD
is an average over all comments, and it can also be lowered by predictions
that are less extreme on every comment, which come closer to the annotators
on contested comments but move away from them on unanimous ones. Identifying
contested comments instead requires a model to give them less extreme
predictions than unanimous ones, which is examined next.

\paragraph{LLM output formats.} For GPT-5.6 Terra, the soft-label prompt gave a
lower MD than the hard-label prompt on all four test sets (0.359--0.487 against
0.412--0.568), and also a higher Macro-F1. For the Arabic LLMs, the soft-label prompt lowered Macro-F1 in 15 of the 16
combinations of model and test set, comparing Fanar on the mixed-dialect test
set only on the comments with a valid reply.
Jais-2 obtained an MD above the constant-median reference on every test set,
up to 1.234 (Egyptian), ALLaM on two of them and Fanar on one. Some
replies were also outside the requested format: on the mixed-dialect test set,
Jais-2 answered 3 for almost every comment, and 958 of the 1,347 replies of
Fanar were numbers above 3 and were excluded. With the perspectivist prompt,
the annotator error rate of the Arabic LLMs was higher than with the
hard-label prompt in nine of the 12 combinations of model and test set; for
GPT-5.6 Terra it was lower on the Saudi and Egyptian test sets and higher on
the Moroccan test set (0.237 against 0.206).

\paragraph{Prompt language and examples.} We also investigate, in Appendix~\ref{app:prompt-effects}, the effect of the prompt language and of few-shot examples. Adding few-shot examples (8, or 24
in the mixed-dialect setting) increased the hard-label Macro-F1 of the LLMs in
35 of the 40 combinations of model, prompt language and test set, by 0.040 on
average, and the English prompt exceeded the Arabic one in 34 of the 40, also
by 0.040 on average, with the largest difference on the Egyptian test set
(0.056). Four of the six exceptions to the English advantage came from GPT-5.6
Terra, by at most 0.006.

\subsection{Identifying contested comments}
\label{sec:res-disagreement}

\begin{table}[htbp]
\centering
\small
\setlength{\tabcolsep}{3.5pt}
\begin{tabular}{llcccc}
\toprule
\textbf{Model} & \textbf{Paradigm} & \textbf{SA} & \textbf{EG} & \textbf{MA} & \textbf{Mix} \\
\midrule
\multicolumn{6}{l}{\textit{Disagreement F1 (predicted contested share $\hat{p}$)}} \\
\textit{All contested} & -- & 0.506 (0.34) & 0.399 (0.25) & 0.498 (0.33) & 0.470 (0.31) \\
MARBERT & Soft   & 0.500 (0.50) & 0.385 (0.35) & 0.453 (0.39) & \underline{0.489} (0.49) \\
        & Persp. & 0.405 (0.28) & 0.335 (0.21) & 0.422 (0.33) & 0.423 (0.29) \\
AraBERT & Soft   & \underline{0.513} (0.45) & \underline{0.416} (0.56) & 0.486 (0.41) & 0.455 (0.37) \\
        & Persp. & 0.401 (0.26) & 0.354 (0.23) & 0.419 (0.29) & 0.356 (0.21) \\
GPT-5.6 Terra & Soft   & 0.407 (0.22) & 0.290 (0.22) & 0.383 (0.23) & 0.400 (0.27) \\
              & Persp. & 0.439 (0.33) & 0.276 (0.22) & 0.336 (0.19) & n/a \\
\midrule
\multicolumn{6}{l}{\textit{Disagreement AUROC}} \\
\textit{Chance} & -- & 0.500 & 0.500 & 0.500 & 0.500 \\
TF-IDF + LR & Hard & 0.586 & 0.592 & 0.596 & 0.600 \\
MARBERT & Hard   & 0.642 & 0.610 & 0.603 & 0.602 \\
        & Soft   & 0.642 & 0.615 & 0.608 & 0.641 \\
        & Persp. & 0.622 & 0.632 & 0.608 & 0.621 \\
AraBERT & Hard   & 0.652 & 0.580 & 0.644 & 0.615 \\
        & Soft   & 0.664 & 0.604 & 0.650 & 0.647 \\
        & Persp. & 0.672 & 0.609 & 0.627 & 0.632 \\
GPT-5.6 Terra & Soft   & 0.620 & 0.537 & 0.596 & 0.601 \\
& Persp. & 0.608 & 0.536 & 0.589 & n/a \\

\bottomrule
\end{tabular}
\caption{Identification of contested comments. Top: Disagreement F1, with
the predicted contested share $\hat{p}$ in parentheses; the reference row
marks every comment as contested and gives the observed contested share.
Scores above this reference are underlined. Bottom: Disagreement AUROC,
using $\hat{s}(1-\hat{s})$ as the score; for GPT-5.6 Terra, $\hat{s}$ is
averaged over the three runs, and the hard-label prompt is not shown, since
the spread of its three runs reflects sampling variation rather than predicted
disagreement.}
\label{tab:res-dis}
\end{table}

Table~\ref{tab:res-dis} reports how well the models identified contested
comments. Three of the 16 encoder scores in Disagreement F1 exceeded the
all-contested reference, all from soft-label models, and in each case the
predicted contested share was higher than the observed share, up to 56\%
against 25\% for AraBERT on the Egyptian test set. The perspectivist
encoders marked 21--33\% of comments as contested, and their Disagreement F1
was below the reference on every test set. GPT-5.6 Terra remained below the
reference on every test set, with predicted contested shares of 19--33\%.

Disagreement AUROC ranged from 0.580 to 0.672 for the Arabic encoders,
against 0.586--0.600 for the TF-IDF baseline. The soft-label encoders reached
a higher AUROC than the corresponding hard-label encoders in seven of the
eight combinations, by 0.005--0.039, and the same value in the eighth. For
GPT-5.6 Terra, it ranged from 0.536 to 0.620 with both prompts, at or below
the level of the TF-IDF baseline on most test sets.

These results partly follow from what each paradigm sees during training.
The soft-label models are trained on the proportion of sexist votes, so
contested comments are marked explicitly in their targets. The
perspectivist models see the same three labels through a shared encoder,
but each head is trained only on the labels of its own annotator, so
disagreement is represented only as differences between heads; in
practice, the heads tended to agree, marking 21--33\% of comments as
contested. The hard-label models see only the majority label.
Accordingly, the soft-label encoders obtained a higher Disagreement F1 than
the perspectivist ones in all eight combinations, although partly by
marking more comments as contested than the annotators did, while their
AUROC was higher in four of the eight, lower in three and equal in one.

Three conclusions follow. First, none of the models reliably identified
contested comments: Disagreement F1 exceeded the all-contested reference
only when a model over-predicted disagreement, and the best AUROC (0.672)
was less than 0.09 above that of a bag-of-words baseline, so whether a
comment is contested is only weakly reflected in its wording as these
models represent it. Second, the training signal matters more than the
model: exposing the encoders to the proportion of votes improved their
ranking of contested comments over the hard-label models in seven of eight
cases, while splitting the votes across annotator heads did not
consistently do so. Third, the uncertainty of an LLM is not a proxy for
human disagreement: GPT-5.6 Terra, whose vote counts and runs varied, did
not rank contested comments above unanimous ones better than the TF-IDF
baseline on most test sets. Together with the lower MD of the
disagreement-aware models (Section~\ref{sec:res-main}), this indicates that
these models came closer to the annotators on average without learning
which comments people disagree on, so flagging contested content for human
review remains an open problem.

\subsection{Contested and unanimous comments}
\label{sec:res-contested}

The hard-label encoders and GPT-5.6 Terra were wrong on 30--45\% of the
contested comments, against 9--23\% of the unanimous ones. The difference
ranged from 0.09 to 0.28, and its 95\% confidence interval excluded zero for
all three models on every test set. Figure~\ref{fig:res-curves} shows
the share of comments that each model labelled sexist at each level of
agreement, on the mixed-dialect test set. This share increased with the number
of annotators who labelled the comment sexist for every model, from 0.12--0.14
at $s=0$ to 0.78--0.80 at $s=1$ for the encoders, and from 0.18--0.23 to 0.93
for GPT-5.6 Terra. The four Arabic LLMs labelled 27--52\% of the comments with
$s=0$ as sexist. For these models, the difference in error rate between contested and
unanimous comments was smaller on the mixed-dialect test set (0.07--0.13), and
on the Egyptian test set it was close to zero or reversed ($-$0.11 to 0.01),
since their over-prediction produced many errors on comments that all three
annotators labelled not sexist.

\begin{center}
\includegraphics[width=\textwidth,height=0.55\textheight,keepaspectratio]{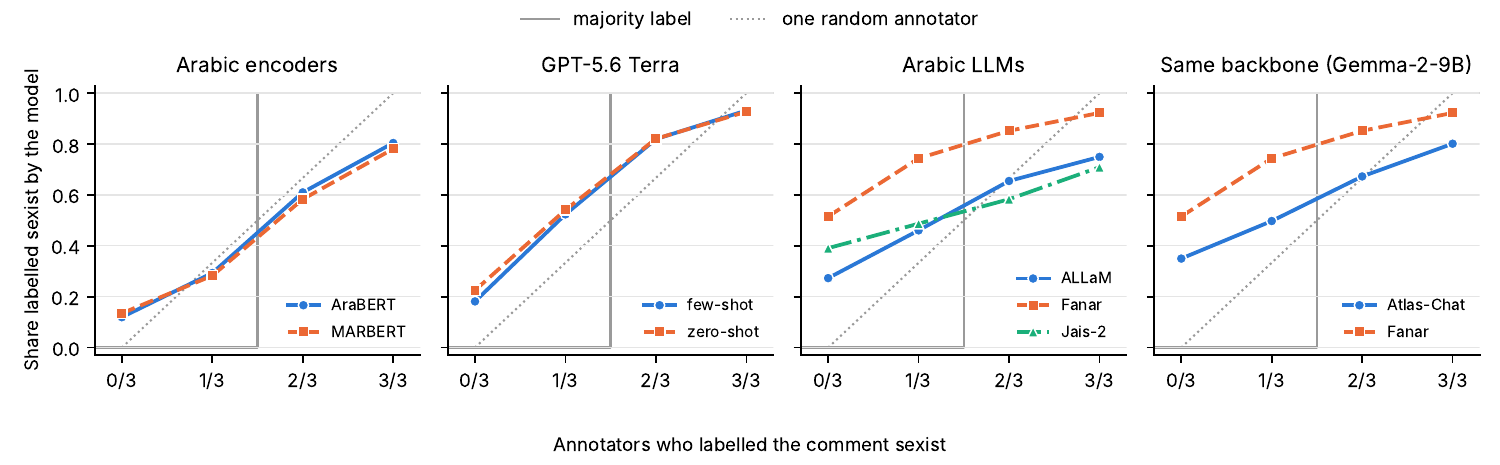}
\captionof{figure}{Share of comments labelled sexist by the hard-label
decision at each level of annotator agreement, on the mixed-dialect test set.
The solid grey line is the majority label; the dotted line is the expected
share for a single annotator drawn at random from the three. Encoder decisions
threshold the mean of the three seeds' predicted probabilities at 0.5; the
LLMs use the English hard-label prompt, few-shot unless marked otherwise.}
\label{fig:res-curves}
\end{center}
 
The two model families differed in where they erred. GPT-5.6 Terra was most
often wrong on comments that only one annotator had labelled sexist
($s=\frac{1}{3}$), where its few-shot prompt predicted sexism in 51--59\% of
cases across the four test sets. The encoders were most often wrong on
comments that two annotators had labelled sexist ($s=\frac{2}{3}$), with
errors of 31--56\%, and they missed 12--26\% of the comments that all three
annotators labelled sexist, against 3--11\% for GPT-5.6 Terra. 

Although none of these models saw the individual votes, the share they
labelled sexist rose gradually with agreement rather than following the
majority label, so the comments on which the annotators disagreed were also
borderline for the models. The figure also shows that the two families
departed from the majority label in opposite directions: the encoders were
conservative, labelling fewer comments sexist than the majority at
$s=\frac{2}{3}$ and $s=1$, whereas GPT-5.6 Terra and, more strongly, the
Arabic LLMs labelled as sexist comments that the majority did not, already
at $s=0$. Errors of similar frequency can therefore lie on opposite sides of
the contested comments, a difference that Macro-F1 does not show and that
Section~\ref{sec:err-direction} examines further.

Figure~\ref{fig:ci-paradigm} breaks the effect of the paradigms down by
agreement. The soft-label encoders lowered MD on the contested comments in
seven of the eight combinations and raised it on the unanimous comments in
six, with intervals excluding zero in all of these cases; the soft-label
MARBERT on the Egyptian test set showed the reverse. 

\begin{center}
\includegraphics[width=\textwidth,height=0.45\textheight,keepaspectratio]{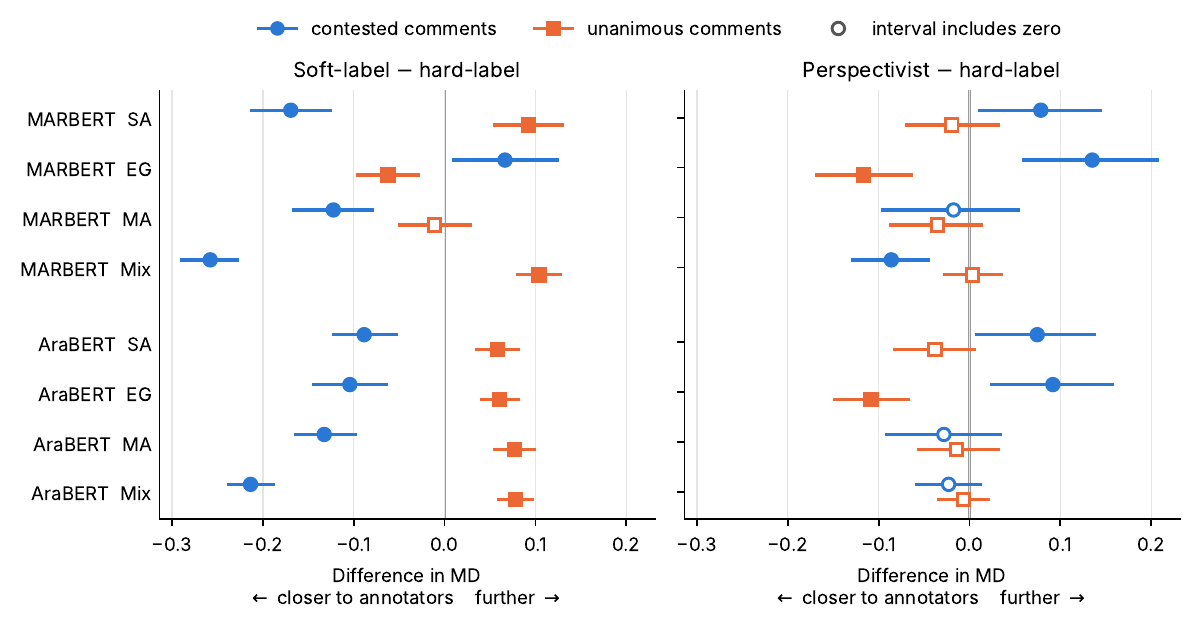}
\captionof{figure}{Difference in MD between each disagreement-aware encoder
and the corresponding hard-label encoder, on the contested and the unanimous
comments, with 95\% bootstrap confidence intervals (2,000 paired resamples,
stratified by dialect). Negative values mean closer alignment with the
annotators. Filled markers: the interval excludes zero. Values are computed
on the seed-averaged predictions.}
\label{fig:ci-paradigm}
\end{center}

The effect of the perspectivist paradigm depended on the setting: in the
in-dialect setting, it lowered MD on the unanimous comments, with an interval
excluding zero only on the Egyptian test set, and raised it on the contested
comments of the Saudi and Egyptian test sets, whereas in the mixed-dialect
setting it lowered MD on the contested comments for MARBERT only, with no
change on the unanimous comments whose interval excluded zero. For the
LLMs, the soft-label prompt lowered MD on the contested comments relative to
the hard-label prompt in 15 of the 16 Arabic LLM--test set combinations, and raised it on the
unanimous comments in 15 of the 16, comparing the two prompts on the comments
with a valid reply to both.

These results qualify the averages in Section~\ref{sec:res-main}. The
soft-label paradigm changed the predictions more than the perspectivist one,
but in opposite directions on the two groups of comments: closer to the
annotators on the contested comments and further from them on the unanimous
ones, consistent with predictions that move away from 0 and 1. For the
encoders, the gain outweighed the loss: averaged over the two encoders, the
soft-label models obtained a lower MD than the hard-label ones on all four
test sets (Table~\ref{tab:res-main}). The perspectivist encoders, whose
changes were smaller and depended on the setting, nevertheless obtained the
lowest mean MD on all four. For the LLMs, the outcome depended on the model:
the soft-label prompt lowered MD for GPT-5.6 Terra on every test set, but
the mean over the five LLMs was higher than with the hard-label prompt on
every test set. Learning from disagreement therefore did not make the models
uniformly closer to the annotators: the soft-label paradigm mainly shifted
where they aligned, towards the contested comments. This makes it the better
choice when contested comments are of interest, for example when flagging
them for review (Section~\ref{sec:res-disagreement}), but not necessarily
when overall alignment is the goal.

\section{Error Analysis}
\label{sec:error-analysis}

This section examines the errors of the hard-label models in more detail:
their direction, the kinds of comments on which they occur, and why
performance was lower on Egyptian Arabic. Throughout, each encoder's decision
is obtained by thresholding the mean of its three seeds' predicted
probabilities at 0.5, and the LLMs use the English few-shot hard-label prompt. Sections~\ref{sec:err-direction} and~\ref{sec:err-qualitative} complete the
answer to RQ1, and Section~\ref{sec:err-egyptian} addresses RQ3.

\subsection{Direction of errors}
\label{sec:err-direction}

The two model families erred in opposite directions. The encoders predicted
the sexist class about as often as the annotators' majority: the number of
comments they labelled sexist was 0.91--1.10 times the number whose majority
label was sexist, and false positives made up 42--58\% of their errors. The
LLMs predicted the sexist class more often than the majority in 18 of the 20
LLM--test set combinations, with ratios of 1.14--1.49 for GPT-5.6 Terra and up
to 2.26 for Fanar (Egyptian); the exceptions were Jais-2 on the Saudi and
Moroccan test sets (0.73 and 0.91). False positives made up 71--86\% of the
errors of GPT-5.6 Terra.
 
The differences in model predictions were also reflected in their agreement
with individual annotators. On the Moroccan test set, where annotators
differed most in their rates of assigning the sexist label, GPT-5.6 Terra
agreed more often with the two annotators who labelled 53--54\% of comments
sexist (83--85\% agreement) than with the annotator who labelled 35\% sexist
(73\% agreement). In contrast, agreement between the encoders and the three
Moroccan annotators varied by at most three percentage points
(Appendix~\ref{app:annotators}). This pattern is consistent with the tendency
of GPT-5.6 Terra to over-predict sexism, although raw agreement alone does not
establish alignment with an annotator's perspective.

This over-prediction was not caused by the definition of sexism alone: the
prompts used the annotators' definition, and with a gender-neutral definition
that does not mention women, GPT-5.6 Terra produced more false positives on
the mixed-dialect test set (Appendix~\ref{app:definition}). A broader
definition therefore adds to the over-prediction, but aligning the
definitions does not remove it.

Errors of both kinds fell disproportionately on contested comments.
Contested comments formed 16--31\% of the comments whose majority label was
not sexist, but 18--61\% of the false positives, and 37--41\% of those whose
majority label was sexist, but 32--74\% of the false negatives. The share of
false positives on contested comments exceeded this reference in all 28
model--test set combinations, and the share of false negatives in 27 of the
28.

\subsection{Qualitative analysis}
\label{sec:err-qualitative}

To examine what lies behind these errors, we coded the errors of the four
strongest systems on the mixed-dialect test set: the hard-label AraBERT and
MARBERT trained on the pooled data, and GPT-5.6 Terra with the English
zero-shot and few-shot hard-label prompts, each taken as the majority vote
over its three runs. The comments were grouped by which systems misclassified
them, and 245 errors were coded: all 48 comments misclassified by all four
systems, the 89 false positives of both GPT-5.6 Terra prompts that both
encoders classified correctly, the 73 false negatives of both encoders that
both GPT-5.6 Terra prompts classified correctly, and the two smaller reverse
groups (11 and 24 comments). Each comment was assigned one category. The categories were defined in two steps: we started from the categories of the EDOS taxonomy
\citep{kirk2023semeval} that applied to the comments, such as derogation,
gender stereotypes and support for the mistreatment of women, and added
further categories for patterns that recurred across the errors but are not
covered by that taxonomy, such as religious or modesty prescriptions,
neutral debate on gender roles and counter-speech. 
The coding was carried out by a single coder. Since only misclassified
comments were coded, the counts show which kinds of comments the errors fell
on, not how often comments of each kind were misclassified.

Table~\ref{tab:err-codes} reports the categories, and
Table~\ref{tab:err-examples} gives examples.

\begin{table}[htbp]
\centering
\small
\setlength{\tabcolsep}{4pt}
\begin{tabular}{llcccc}
\toprule
\textbf{Category} & \textbf{Majority} & \textbf{All four} & \textbf{GPT FP} & \textbf{Encoder FN} & \textbf{Total} \\
\midrule
Religious or modesty prescription            & not sexist & 2  & 35 & -- & 38 \\
Neutral debate on roles, money or law        & not sexist & 4  & 22 & -- & 31 \\
Attack on one specific woman                 & not sexist & 11 & 12 & -- & 26 \\
Derogatory generalisation about women        & not sexist & 11 & 8  & -- & 19 \\
Counter-speech                               & not sexist & 2  & 5  & -- & 14 \\
Criticism of men                             & not sexist & 1  & 1  & -- & 5  \\
\midrule
Generalisation within a longer narrative     & sexist     & -- & -- & 21 & 21 \\
Prescriptive or benevolent gender roles      & sexist     & -- & -- & 20 & 20 \\
Sarcasm, metaphor or insult to one woman     & sexist     & 5  & -- & 9  & 18 \\
Victim-blaming                               & sexist     & 2  & -- & 9  & 12 \\
Dependent on video or thread context         & sexist     & 6  & -- & 3  & 11 \\
Modesty policing                             & sexist     & -- & -- & 8  & 8  \\
Opposition to feminism                       & sexist     & -- & -- & 3  & 7  \\
\bottomrule
\end{tabular}
\caption{Categories of the 245 coded errors on the mixed-dialect test set.
\textit{Majority}: majority label of the comments in each category.
\textit{All four}: misclassified by AraBERT, MARBERT and both GPT-5.6 Terra
prompts; \textit{GPT FP}: false positives of both GPT-5.6 Terra prompts only;
\textit{Encoder FN}: false negatives of both encoders only. \textit{Total}
also includes the two smaller reverse groups. Not shown: other (11
comments), opposition to feminism with a majority label of not sexist (3)
and a doubtful positive (1).}
\label{tab:err-codes}
\end{table}

\paragraph{False positives.}
Among the 89 false positives of GPT-5.6 Terra only, the most frequent
categories were religious or modesty prescription (35), neutral debate on
gender roles, money or law (22) and attacks on one specific woman (12).
Criticism of men, one of the most frequent categories with the
gender-neutral definition, accounted for only one of these false positives
(Appendix~\ref{app:definition}). Of the 35 religious or modesty
prescriptions, 23 had been labelled not sexist by all three annotators.
These comments typically prescribe dress or behaviour for women on
religious grounds. The annotators mostly read them as religious advice
rather than as sexism, and the encoders, trained on their labels, classified
them correctly, while the annotators did label as sexist the comments that
policed women's modesty (Table~\ref{tab:err-codes}). Whether the errors of
GPT-5.6 Terra reflect a bias towards religious content or a reading of
gender norms that differs from that of the annotators cannot be determined
from these data, but this is a boundary on which the LLM and the annotators
diverged.

Among the 48 comments misclassified by all four systems, the most frequent
categories were attacks on one specific woman (11) and derogatory
generalisations about women whose majority label was not sexist (11). Of the 19 comments in the latter category, 14 had been labelled sexist by one of
the three annotators.

\paragraph{False negatives.}
Among the 73 false negatives of the encoders only, the most frequent
categories were sexist generalisations within a longer comment (21) and
prescriptive or benevolent statements about gender roles (20), followed by
victim-blaming and sarcasm, metaphor or insults to one woman (9 each); 35 of
these 73 comments had been labelled sexist by all three annotators. Among the
14 false negatives of all four systems, 6 were comments whose sexist reading
depended on the video or the surrounding thread.

\begin{table}[htbp]
\centering
\small
\begin{tabular}{p{0.46\textwidth}p{0.30\textwidth}l}
\toprule
\textbf{Comment (translation)} & \textbf{Category} & \textbf{Error} \\
\midrule
\textarabic{يا اختي الحجاب فرض علينا كمسلمات مش اختياري}
(``Sister, the hijab is an obligation for us as Muslim women, not a
choice''; Egyptian, $s = 0$)
& Religious or modesty prescription & GPT FP \\
\addlinespace
\textarabic{لازم احد يساعد في البيت اذا كانت الزوجه تكرف دايم وتدرس متى تفضى لزوجها ؟}
(``Someone has to help at home if the wife is always working hard and
studying; when will she have time for her husband?''; Saudi, $s = 0$)
& Neutral debate on roles, money or law & GPT FP \\
\addlinespace
\textarabic{صح واجب الرجل و واجب الزوجة الطاعة وتربية الابناء}
(``Right, the man's duty, and the wife's duty is obedience and raising the
children''; Saudi, $s = \tfrac{2}{3}$)
& Prescriptive or benevolent gender roles & Encoder FN \\
\addlinespace
\textarabic{اقسم بالله كلامك صح انا صيدلاني وغالبية الادوية النفسية تستهلكها العوانس والمطلقات}
(``I swear you are right; I am a pharmacist, and most psychiatric medication
is taken by spinsters and divorced women''; Egyptian, $s = 1$)
& Generalisation within a longer comment & Encoder FN \\
\bottomrule
\end{tabular}
\caption{Examples of coded errors, with the dialect and the proportion $s$
of annotators who labelled the comment sexist.}
\label{tab:err-examples}
\end{table}

\subsection{Lower performance on Egyptian Arabic}
\label{sec:err-egyptian}

Most models obtained their lowest in-dialect score on the Egyptian test set
(Table~\ref{tab:res-main}). We examined two explanations for this gap: the
composition of the Egyptian test set, and its level of dialectness.

\paragraph{Composition.}
The Egyptian test set has the lowest proportion of sexist comments (34.4\%)
and of contested comments (24.9\%). To estimate how much of the gap this
composition accounts for, we computed each model's error rate at the four
agreement levels, pooled over the three dialect test sets, and applied these
rates to the number of comments at each level in each test set. The resulting
expected Macro-F1 differs between test sets only through their composition.
Figure~\ref{fig:ci-egyptian} compares the observed and expected gap of the
Egyptian test set, defined as its Macro-F1 minus the mean of the Saudi and Moroccan values. 

\begin{center}
\includegraphics[width=0.65\textwidth,height=0.45\textheight,keepaspectratio]{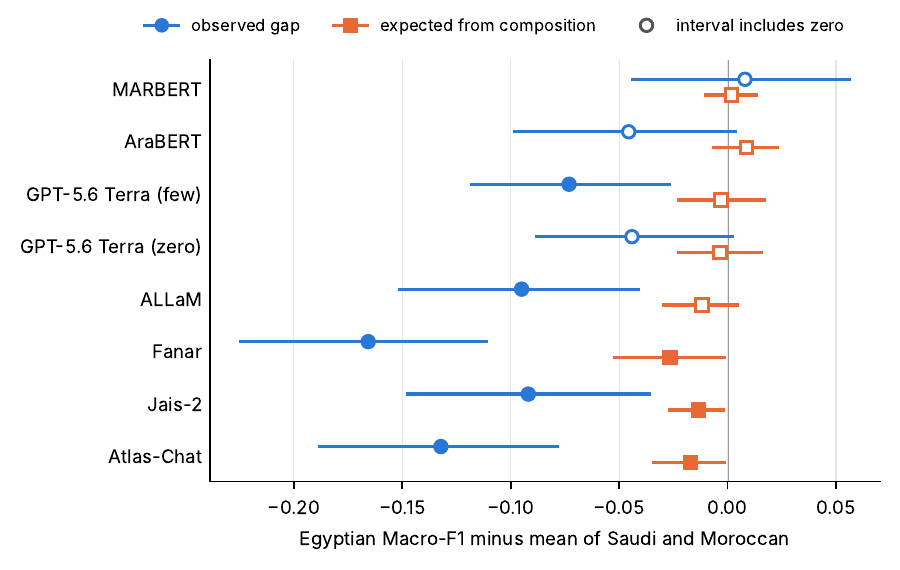}
\captionof{figure}{Gap of the Egyptian test set (Egyptian Macro-F1 minus the mean of the Saudi and Moroccan values), observed and expected from the composition of
the test sets, with 95\% bootstrap confidence intervals (2,000 resamples of
each dialect test set). Filled markers: the interval excludes zero.
GPT-5.6 Terra is shown with both the few-shot and the zero-shot prompt.}
\label{fig:ci-egyptian}
\end{center}

The expected gap ranged from $-$0.027 to $+$0.009; its 95\% confidence interval
included zero for the encoders, GPT-5.6 Terra and ALLaM, and excluded it for
Fanar, Jais-2 and Atlas-Chat, by at most 0.027. The observed gap had an interval
excluding zero for GPT-5.6 Terra with the few-shot prompt ($-$0.073) and for
the four Arabic LLMs ($-$0.092 to $-$0.166); for the encoders and for GPT-5.6
Terra with the zero-shot prompt ($-$0.044), the interval included zero. The
difference between the observed and expected gaps, which reflects differences
in error rate within each agreement level, had an interval excluding zero for
AraBERT, GPT-5.6 Terra with the few-shot prompt and the four Arabic LLMs.

In other words, the composition of the Egyptian test set accounts for only
a small part of the gap: for these models, the Egyptian comments were
misclassified more often than Saudi and Moroccan comments at the same level
of agreement. The gap therefore has another source, and we next examine whether it is
the dialectness of the comments.

\paragraph{Dialectness.}
We estimated the dialectness of each test comment with ALDi
\citep{keleg2023aldi}, which assigns a score from 0 (Modern Standard Arabic)
to 1 (fully dialectal). The Egyptian comments had the highest mean score
(0.747, against 0.600 for Moroccan and 0.485 for Saudi). Within each dialect,
the score was not associated with whether a comment was contested (the
point-biserial correlation was $-0.03$ to $0.09$ and not significant on any
test set). To compare comments that differ in dialectness within the same
dialect, each test set was divided into ALDi quartiles. All models had their
highest error rate in the most dialectal quartile, but the increase from the
least to the most dialectal quartile had a 95\% confidence interval excluding
zero only for two of the four Arabic LLMs (ALLaM 0.125 and Fanar 0.191),
and not for the encoders (0.004--0.036) or GPT-5.6 Terra (0.007). 

Neither the composition of the Egyptian test set nor the dialectness of its
comments therefore explains most of the gap, which lay in how the models
responded to Egyptian comments at the same level of agreement. Possible reasons, namely the comments themselves, the models' competence in Egyptian
Arabic and how the Egyptian annotators judged sexism, are discussed in
Section~\ref{sec:disc-egyptian}.

\section{Discussion}
\label{sec:discussion}

\subsection{Model errors and annotator disagreement (RQ1)}
\label{sec:disc-error}

\paragraph{\textbf{Answering RQ1:}} \textit{How are the errors of these models distributed across levels of
annotator agreement, and do models differ in the kinds of errors they make
beyond what Macro-F1 captures?}

The errors of all models concentrated on the comments on
which the annotators disagreed, consistent with earlier findings that
agreement levels strongly affect classifier performance
\citep{leonardelli2021agreeing}, and models with similar Macro-F1 made
different kinds of errors. On a
contested comment, a randomly chosen annotator disagrees with the majority
label with a probability of one in three, and the error of the models on
these comments was of the same order (30--45\%). Much of what is counted as
error there therefore coincides with genuine disagreement among the
annotators: the majority label records which side had two votes rather than
a fact that a model could be expected to recover.

The two model families nonetheless reached a similar Macro-F1 on the
mixed-dialect test set through opposite errors: 0.799 and 0.771 on average.
The difference was small, with a 95\% confidence interval of [0.003, 0.053]
on the means over runs and seeds and of [$-$0.003, 0.052] on the aggregated
decisions. GPT-5.6 Terra over-predicted the sexist
class, most often on comments that only one annotator had labelled sexist, in
line with the tendency of LLMs to label ambiguous content as offensive
\citep{lu-etal-2025-llm}. Many of its false positives, however, fell on
comments that all three annotators had labelled not sexist, such as religious
prescriptions on modesty and neutral debate on gender roles, even with the
annotators' definition of sexism in the prompt
(Appendix~\ref{app:definition}). The encoders, in contrast, missed sexism without explicit derogation, such as prescriptive or benevolent
statements about gender roles \citep{glick1996ambivalent} and generalisations
embedded in longer comments, which lack the lexical cues a classifier trained
on about 1,500 comments per dialect may rely on.

One category illustrates why the agreement level should be kept alongside
the majority label. Of the 19 coded comments that made derogatory
generalisations about women but had a majority label of not sexist, 14 had
been labelled sexist by one annotator. A model that labels such a comment
sexist agrees with that annotator, yet a hard-label evaluation counts this
agreement as an error.

\subsection{What disagreement-aware learning changes (RQ2)}
\label{sec:disc-dal}

\paragraph{\textbf{Answering RQ2:}} \textit{Does learning from soft labels or from individual annotators
change how well model predictions align with the annotators, and on which comments?}

Learning from soft labels or from individual annotations changed where the predictions aligned with the annotators more than how
closely they aligned overall. Both paradigms preserved Macro-F1:
the confidence interval of the difference from the hard-label paradigm
included zero in all 16 comparisons.
Their benefit, however, lies less in overall alignment, which changed reliably
in five of 16 comparisons, than in \emph{where} the predictions are close to
the annotators.
Previous work has suggested that soft labels help most on
ambiguous instances \citep{fornaciari2021beyond}, and that multi-annotator
models capture minority labels where majority-label models are uncertain
\citep{anand2024don}. Our results add the cost side of this picture: the
soft-label encoders improved alignment on contested comments at the expense of
unanimous ones, and marked more comments as contested than the annotators did.
This is consistent with a general shift of predictions towards the middle of
the scale, which brings a model closer to the annotators where they disagree
and further from them where they agree. The effect of the perspectivist
paradigm depended on the setting and was clear mainly on the Egyptian test
set.

Neither paradigm reliably identified the contested comments. The soft-label
encoders ranked contested comments slightly above unanimous ones, but their
Disagreement AUROC exceeded that of a TF-IDF baseline by only 0.01--0.08.
A model trained to predict sexism thus carries only a weak signal of where
annotators disagree, and dedicated predictors of disagreement
\citep{yelkanat2026decomposing} may be needed for this purpose.

Prompting LLMs for a vote count or for individual annotators' judgements did
not close this gap: the perspectivist prompts did not consistently reproduce
the annotators better than the hard-label prompt, and for the smaller Arabic
LLMs, a vote count was a harder output to produce than a binary decision.
The two perspectivist settings also represent annotators differently: the
encoders learn a separate classification head from each annotator's labels,
whereas the LLMs are asked to imitate an annotator from a small number of
that annotator's labelled examples. The lack of improvement from the
perspectivist prompts should therefore not be read as evidence against the
perspectivist approach in general.

\subsection{Lower performance on Egyptian Arabic (RQ3)}
\label{sec:disc-egyptian}

\paragraph{\textbf{Answering RQ3:}} \textit{Does model performance differ across the three dialects and, if
so, to what extent are these differences related to the composition of each
test set (its proportion of sexist and of contested comments) and to the
dialectness of its comments?}

Performance differed across the dialects and was lowest on
the Egyptian test set for every LLM. This difference was not explained by
the composition of the test set, which accounted for at most 0.027 of gaps
of 0.07--0.17, or by the dialectness of its comments within the dialect. Most of the gap lay in how
the models responded to Egyptian comments at the same agreement level. This
could reflect the comments themselves, such as their topics, or the models'
competence in Egyptian Arabic, and the two cannot be separated with the
present data. A third possibility is that the Egyptian annotators drew the
line between sexist and not sexist differently from the annotators of the
other dialects: they labelled the fewest comments sexist and agreed with each
other most often. Since each dialect was labelled by its own annotators,
annotator group and dialect are confounded in this design. MARBERT, pretrained on dialectal Arabic social media \citep{abdul2021arbert},
was the only model whose Egyptian score was not lower than the mean of the
other two dialects (observed gap 0.008, with an interval including zero), but
a single model does not establish a cause.

\subsection{Implications for content moderation}
\label{sec:disc-implications}

We draw four main implications from the results for the automatic moderation of sexist
content in Arabic.

First, models with similar Macro-F1 can fail in opposite
ways, with different consequences for users: a model like GPT-5.6 Terra could incorrectly flag  religious discussion of modesty and neutral debate on gender roles that annotators did not consider sexist, while the fine-tuned encoders could miss prescriptive and benevolent sexism. Evaluations should therefore be more detailed, including
reporting the direction of errors and their distribution across agreement
levels, not only an aggregate score.

Second, datasets should preserve and share individual annotations, since the majority
label hides the comments on which judgements differ and turns agreement with
a minority annotator into an error.

Third, results should be reported by
dialect, since pooled scores hide differences between dialects, and the
definition of sexism given to an LLM should match the one used by
the annotators, since a broader definition increased false positives
(Appendix~\ref{app:definition}).

Fourth, an Arabic prompt is not necessarily the better choice for Arabic
content with current models: the English prompt gave a higher hard-label Macro-F1 in 34 of 40
settings, including for the Arabic-centric LLMs, and the six exceptions were
small (Section~\ref{sec:res-main}). Nor did adaptation to a dialect give a
clear advantage on that dialect: Atlas-Chat, adapted to Moroccan Arabic,
outperformed Fanar, which shares its backbone, on all three test sets, but its
advantage was largest on the Moroccan test set in only two of the four prompt
conditions.

\section{Conclusion}
\label{sec:conclusion}

In this work, we investigate what majority-label evaluation hides in Arabic
sexism detection, and how fine-tuned encoders and LLMs behave when the
annotators of a comment disagree. To support this investigation, we developed MuDAS, a dataset of YouTube comments in Saudi, Egyptian and
Moroccan Arabic, each labelled for sexism by three native speakers of its
dialect with the individual annotations preserved, and used it to compare
hard-label, soft-label and perspectivist approaches across both model
families. Model errors concentrated on the comments on
which annotators disagreed, and models with similar Macro-F1 erred in
opposite directions: GPT-5.6 Terra over-flagged gender-related content even with the annotators' definition of sexism in its prompt, while fine-tuned Arabic encoders missed implicit and
prescriptive sexism.

Disagreement-aware learning preserved Macro-F1 but shifted where the
predictions aligned with the annotators, without identifying the contested comments. Lower performance on Egyptian Arabic was not fully explained by the examined measures of test-set composition and
dialectness, and other explanations remain open. Future work will follow three directions. First, we will analyse
judgements of a subset of comments from each dialect, collected from the annotators of the other two dialects, to test whether sexism judgements
differ across Arabic-speaking communities. Second, we will have a random sample of the Egyptian comments labelled by new Egyptian annotators under the same conditions, to test whether the lower performance on Egyptian
Arabic reflects the judgements of the original annotators rather than the comments themselves. Third, we will train models on one dialect and test them on the others, to measure how well sexism detection transfers across Arabic dialects.

\section{Limitations}
\label{sec:disc-limitations}

While our work advances research on Arabic sexism detection by considering individual annotations and the cultural and linguistic differences across dialects, this is not without limitations.

All comments of a dialect were labelled by the same three annotators, who
cannot represent the range of views within their community. With three
annotators per comment, the proportion $s$ takes
only four values, and the distinction between contested and unanimous
comments rests on a single vote. The label was binary, so the annotations do not show which aspects of a
comment the annotators disagreed on. By design, each comment was judged only by annotators of its own dialect, to
capture how sexism is judged within each community; differences between
dialects therefore cannot be separated from differences between annotator
groups, and judgements of the same comments across dialect communities are
left to future work. 

The test sets contain
430--465 comments per dialect, and some comparisons rest on fewer than 100
comments. Each paradigm was selected on its own validation criterion, so differences between paradigms reflect both the training objective and the selection criterion. All comments come from one platform and from videos selected on gender-related topics. Because comments predicted as sexist by the pre-screening model were
oversampled, the dataset is enriched for likely sexist content and does not
represent the natural prevalence of sexism on YouTube; model performance may
therefore differ at natural prevalence. The split was made at the level of comments, so comments from the same
video, which may share topic and vocabulary, can appear in both the
training and the test sets.

The open-weight LLMs were run once with greedy decoding, the prompts were not
tuned for each model, and only one number of few-shot examples (eight per
dialect) was tested. GPT-5.6 Terra was run three times to account for
variation between runs, but it is a closed model accessed through an API, so
its results may not be exactly reproducible if the model is updated. The prompts gave the definition of sexism from the
guidelines but not the boundary cases illustrated in them (Appendix~\ref{app:guideline-examples}). 
Some replies of the Arabic LLMs to the soft-label prompt were outside the
requested range and were excluded, most notably 958 of the 1,347 replies of
Fanar on the mixed-dialect test set. No model received the description of the
video, although some comments are sexist only in its context. The
perspectivist models were evaluated on the annotators who labelled the
training data, and their ability to generalise to new annotators was not
tested.

Finally, the qualitative coding was carried out by a single coder and covered
only misclassified comments, so the error rate of each category is unknown;
the bootstrap intervals for the encoders were computed on the seed-averaged
predictions and do not include the variation between seeds.

\section*{Ethics statement}
The study was approved by the research ethics committee of Queen Mary
University of London (ref.\ DSEECS26.020). All annotators gave informed
consent, were informed that the task involved offensive content, and could
withdraw at any time. They were paid \pounds15 per hour, above
Prolific's recommended rate. User mentions were replaced with \texttt{@USER}
to protect the privacy of YouTube users.
\section*{Funding}
This work was supported by the Royal Embassy of Saudi Arabia Cultural
Bureau. The funder had no role in the study design, the collection, analysis
and interpretation of the data, the writing of the article, or the decision
to submit it for publication.

\section*{Declaration of competing interest}
Given his role as Area Editor, Arkaitz Zubiaga had no involvement in the
peer review of this article and has no access to information regarding its
peer review. Full responsibility for the editorial process for this article
was delegated to another journal editor. The authors declare that they have
no other known competing financial interests or personal relationships that
could have appeared to influence the work reported in this paper.

\section*{Data availability}
MuDAS is available under gated access for research purposes at
\url{https://huggingface.co/datasets/alaabaz/MuDAS}; access is granted on
request.

\section*{Declaration of generative AI and AI-assisted technologies in the manuscript preparation process}
During the preparation of this work, the authors used Claude (Anthropic) in
order to improve the language and readability of the manuscript and to assist
with LaTeX formatting and analysis code. After using this tool, the authors
reviewed and edited the content as needed and take full responsibility for
the content of the publication.

\printcredits

\bibliographystyle{cas-model2-names}

\bibliography{cas-refs}

\appendix
\appendix
\counterwithin{table}{section}
\counterwithin{figure}{section}

\section{Search keywords}
\label{app:keywords}

Table~\ref{tab:app-keywords} lists the 32 keywords used to search for videos
on YouTube (Section~\ref{sec:ds-collection}). The keywords were searched in
Arabic as written; the English translations are given for reference. Not
every keyword yielded a video that met the selection criteria, so videos
were recorded by theme rather than by keyword
(Table~\ref{tab:app-videos}). All five themes are represented in each
dialect, although unevenly: roles and relationships and marriage and divorce
account for most videos in every dialect.

\begin{table}[!ht]
\centering

\begin{tabular}{lcccc}
\toprule
\textbf{Theme} & \textbf{Saudi} & \textbf{Egyptian} & \textbf{Moroccan} & \textbf{Total} \\
\midrule
Marriage and divorce     &  7 &  5 &  7 & 19 \\
Work and independence    &  2 &  1 &  2 &  5 \\
Appearance and behaviour &  3 &  4 &  4 & 11 \\
Roles and relationships  & 12 & 10 &  5 & 27 \\
Controversial issues     &  1 &  4 &  1 &  6 \\
\midrule
Total                    & 25 & 24 & 19 & 68 \\
\bottomrule
\end{tabular}
\caption{Number of selected videos by theme and dialect.}
\label{tab:app-videos}
\end{table}

\begin{table}[htbp]
\centering
\small
\setlength{\tabcolsep}{5pt}
\begin{tabular}{llp{6.2cm}}
\toprule
\textbf{Theme} & \textbf{Arabic keyword} & \textbf{English translation} \\
\midrule
Marriage and divorce
 & \textarabic{العزوف عن الزواج / عنوسة} & Avoidance of marriage / spinsterhood \\
 & \textarabic{الخلع وحقوق المرأة في الطلاق} & \textit{Khul'} (divorce initiated by the wife) and women's rights in divorce \\
 & \textarabic{النفقة وحقوق المطلقة} & Alimony and the rights of divorced women \\
 & \textarabic{تعدد الزوجات} & Polygamy \\
 & \textarabic{الولاية في الزواج} & Male guardianship in marriage \\
 & \textarabic{فارق السن في الزواج} & Age gap in marriage \\
 & \textarabic{المهر والغلاء} & Dowry and rising marriage costs \\
 & \textarabic{زواج المسيار} & \textit{Misyar} marriage (spouses waive some rights, e.g.\ cohabitation) \\
 & \textarabic{الزواج من أجنبية/أجنبي} & Marriage to a foreign woman or man \\
\addlinespace
Work and independence
 & \textarabic{عمل المرأة / المرأة العاملة} & Women's employment / working women \\
 & \textarabic{سواقة المرأة} & Women driving \\
 & \textarabic{المرأة في المناصب القيادية} & Women in leadership positions \\
 & \textarabic{الاختلاط في العمل} & Gender mixing in the workplace \\
 & \textarabic{سفر المرأة بدون محرم} & Women travelling without a male guardian (\textit{mahram}) \\
 & \textarabic{الاستقلال المادي للمرأة} & Women's financial independence \\
\addlinespace
Appearance and behaviour
 & \textarabic{الحجاب / خلع الحجاب} & Hijab / removing the hijab \\
 & \textarabic{النقاب} & Niqab (face veil) \\
 & \textarabic{لبس المرأة / الموضة} & Women's clothing / fashion \\
 & \textarabic{المكياج والتبرج} & Makeup and adornment \\
 & \textarabic{صوت المرأة} & Women's voice (in the media or on YouTube) \\
\addlinespace
Roles and relationships
 & \textarabic{النسوية / الفيمنزم} & Feminism \\
 & \textarabic{حقوق المرأة} & Women's rights \\
 & \textarabic{القوامة} & Male authority in the household (\textit{qiwamah}) \\
 & \textarabic{طاعة الزوج} & Obedience to the husband \\
 & \textarabic{المرأة والأمومة مقابل الكاريير} & Motherhood versus career \\
 & \textarabic{نشوز الزوجة} & Wife's disobedience (\textit{nushuz}) \\
 & \textarabic{خروج المرأة من البيت} & Women leaving the home \\
\addlinespace
Controversial issues
 & \textarabic{التحرش} & Sexual harassment \\
 & \textarabic{العنف الأسري} & Domestic violence \\
 & \textarabic{الحضانة بعد الطلاق} & Child custody after divorce \\
 & \textarabic{ميراث المرأة} & Women's inheritance \\
 & \textarabic{شهادة المرأة} & Women's testimony in court \\
\bottomrule
\end{tabular}
\caption{The 32 search keywords used to search for videos, grouped into
five themes.}
\label{tab:app-keywords}
\end{table}

\section{Llama Pre-screening prompt}
\label{app:prescreen}

The following prompt was used with Llama-3.1-8B-Instruct, with greedy
decoding, to pre-screen comments before sampling. Outputs containing neither
label (0.2\% of the comments) were marked as unclear and grouped with the
comments predicted as not sexist during sampling
(Section~\ref{sec:ds-sampling}).

\begin{otherlanguage}{arabic}
صنّف التعليق العربي التالي حسب ما إذا كان يتضمن تحيزًا جنسيًا ضد المرأة.

التحيز الجنسي يشمل الإساءة أو التمييز أو الصور النمطية السلبية تجاه المرأة
بسبب جنسها، سواء بشكل مباشر أو ضمني.

أجب بتصنيف واحد فقط: \texttt{SEXIST} أو \texttt{NOT\_SEXIST}

التعليق: «\{text\}»

التصنيف:
\end{otherlanguage}

\paragraph{English translation.}
\begin{quote}
Classify the following Arabic comment according to whether it contains
sexist bias against women.

Sexist bias includes abuse, discrimination or negative stereotypes towards
women because of their gender, whether direct or implicit.

Answer with a single label only: \texttt{SEXIST} or \texttt{NOT\_SEXIST}

Comment: ``\{text\}''

Label:
\end{quote}

\section{Examples from the annotation guidelines}
\label{app:guideline-examples}

Tables~\ref{tab:app-sexist-examples} and~\ref{tab:app-nonsexist-examples}
give the examples used in the annotation guidelines. The three dialect
versions of each example express the same sentence, so a single English
translation is given.
 
\begin{table}[!ht]
\centering
\footnotesize
\renewcommand{\arraystretch}{1.25}
\begin{tabular}{>{\raggedright\arraybackslash}p{2.9cm}
                >{\raggedleft\arraybackslash}p{2.35cm}
                >{\raggedleft\arraybackslash}p{2.35cm}
                >{\raggedleft\arraybackslash}p{2.35cm}
                >{\raggedright\arraybackslash}p{2.9cm}}
\toprule
\textbf{Type} & \textbf{Saudi} & \textbf{Egyptian} & \textbf{Moroccan} & \textbf{Translation} \\
\midrule
Stereotyping and objectification
 & \textarabic{أغلب الحريم تفكيرهم سطحي}
 & \textarabic{أغلب الستات تفكيرهم سطحي}
 & \textarabic{أغلب العيالات تفكيرهم سطحي}
 & Most women think superficially \\
\addlinespace
Insult and belittling
 & \textarabic{الحريم ما عندهم عقل}
 & \textarabic{الستات مفيش عندهم عقل}
 & \textarabic{العيالات ما عندهمش عقل}
 & Women have no brains \\
\addlinespace
Dominance
 & \textarabic{مكانها البيت}
 & \textarabic{مكانها البيت}
 & \textarabic{بلاصتها فالدار}
 & Her place is in the home \\
\addlinespace
Violence and harassment
 & \textarabic{المفروض أحد يضربها عشان تتأدب}
 & \textarabic{المفروض حد يضربها عشان تتأدب}
 & \textarabic{خاص شي واحد يضربها باش تتأدب}
 & Someone should beat her to teach her manners \\
\addlinespace
Endorsement and justification
 & \textarabic{هو ماله ذنب، هي اللي استفزته}
 & \textarabic{هو ماله ذنب، هي اللي استفزته}
 & \textarabic{هو ما عندوش ذنب، هي اللي حرّكاتو}
 & It is not his fault; she provoked him \\
\addlinespace
Doubt and victim blaming
 & \textarabic{لو تحترمين نفسك ما صار لك شي}
 & \textarabic{لو بتحترمي نفسك مكانش حصل لك حاجة}
 & \textarabic{لو كنتي كتحترمي راسك ما كان وقع ليك والو}
 & If you respected yourself, nothing would have happened to you \\
\addlinespace
Benevolent sexism
 & \textarabic{المرة جوهرة لازم الرجال يصونها ويحميها}
 & \textarabic{الست جوهرة لازم الراجل يصونها ويحميها}
 & \textarabic{المرا جوهرة خاص الراجل يحافظ عليها ويحميها}
 & A woman is a jewel; a man must guard and protect her \\
\addlinespace
Ideological discourse and inequality
 & \textarabic{النسوية خربت البيوت ودمرت الأسرة}
 & \textarabic{النسوية خربت البيوت ودمرت الأسرة}
 & \textarabic{النسوية خرّبات الديار ودمّرات العائلة}
 & Feminism has ruined homes and destroyed the family \\
\bottomrule
\end{tabular}
\caption{Examples of the eight types of sexism in the annotation
guidelines, in each dialect.}
\label{tab:app-sexist-examples}
\end{table}

\begin{table}[htbp]
\centering
\footnotesize
\renewcommand{\arraystretch}{1.25}
\begin{tabular}{>{\raggedright\arraybackslash}p{2.9cm}
                >{\raggedleft\arraybackslash}p{2.35cm}
                >{\raggedleft\arraybackslash}p{2.35cm}
                >{\raggedleft\arraybackslash}p{2.35cm}
                >{\raggedright\arraybackslash}p{2.9cm}}
\toprule
\textbf{Category} & \textbf{Saudi} & \textbf{Egyptian} & \textbf{Moroccan} & \textbf{Translation} \\
\midrule
Praise of women
 & \textarabic{ما شاء الله عليها، امرأة ناجحة}
 & \textarabic{ما شاء الله عليها، ست ناجحة}
 & \textarabic{ما شاء الله عليها، مرا ناجحة}
 & Mashallah, she is a successful woman \\
\addlinespace
Neutral text
 & \textarabic{الوزيرة أعلنت القرار الجديد}
 & \textarabic{الوزيرة أعلنت القرار الجديد}
 & \textarabic{الوزيرة أعلنت القرار الجديد}
 & The (female) minister announced the new decision \\
\addlinespace
Criticism of a specific woman, unrelated to her gender
 & \textarabic{مو متفق مع رأي الوزيرة}
 & \textarabic{مش متفق مع رأي الوزيرة}
 & \textarabic{ماشي متفق مع رأي الوزيرة}
 & I do not agree with the minister's view \\
\addlinespace
A woman's own account of mistreatment
 & \textarabic{أنا أشتغل ونتقاسم المصاريف، ومع ذلك يعاملني بطريقة سيئة}
 & \textarabic{أنا بشتغل وبنتقاسم المصاريف، ومع ذلك بيعاملني بطريقة مهينة}
 & \textarabic{أنا كنخدم وكنقسمو المصاريف، ومع هادشي كيعاملني بطريقة مهينة}
 & I work and we share the expenses, yet he treats me badly \\
\addlinespace
Criticism of discrimination or oppressive practices
 & \textarabic{مو لأنها مو محجبة يعني تستاهل التحرش}
 & \textarabic{مش لأنها مش محجبة يبقى تستاهل التحرش}
 & \textarabic{ماشي حيت ماحجباش يعني كتستاهل التحرش}
 & Not wearing a hijab does not mean she deserves to be harassed \\
\bottomrule
\end{tabular}
\caption{Examples of the five categories of non-sexist comments in the
annotation guidelines, in each dialect.}
\label{tab:app-nonsexist-examples}
\end{table}

\section{LLM prompts}
\label{app:prompts}

This appendix gives the prompts used for the large language models. Each
prompt was written in English and then translated into Arabic. In every
prompt, \texttt{\{comment\}} is replaced by the comment being classified,
and \texttt{\{examples\}} by the few-shot examples drawn from the training
pool.

\subsection*{Hard label, zero-shot}

\textbf{English}

\textit{System:} You are an expert annotator of Arabic social media
content. You label comments for sexism. Answer with one word only.

\textit{User:} Is the following Arabic comment sexist? A comment is sexist if it demeans women, discriminates against them,
reinforces stereotypes about them, or justifies harm done to them, whether
directed at women as a group or at an individual woman because she is a
woman.
Comment: \texttt{\{comment\}}. Answer with exactly one word: Yes or No.

\medskip
\textbf{Arabic}

\begin{otherlanguage}{arabic}
النظام: أنت مُصنِّف خبير لمحتوى وسائل التواصل الاجتماعي باللغة العربية.
تُصنِّف التعليقات من حيث التحيّز الجنسي. أجب بكلمة واحدة فقط.

المستخدم: هل التعليق العربي التالي متحيّز جنسيًا؟ يُعدّ التعليق متحيّزًا جنسيًا إذا كان يحطّ من قيمة المرأة، أو يميّز ضدها، أو
يكرّس صورًا نمطية عنها، أو يبرّر الأذى الواقع عليها، سواء كان موجّهًا للنساء
كفئة أو لامرأة بعينها بسبب كونها امرأة. التعليق: [comment].
أجب بكلمة واحدة فقط: نعم أو لا.
\end{otherlanguage}

\subsection*{Soft label, zero-shot}

\textbf{English}

\textit{System:} You are an expert annotator of Arabic social media
content. Answer with a single digit only.

\textit{User:} Three Arabic-speaking annotators independently judged
whether the following comment is sexist. A comment is sexist if it demeans women, discriminates against them,
reinforces stereotypes about them, or justifies harm done to them, whether
directed at women as a group or at an individual woman because she is a
woman. Comment:
\texttt{\{comment\}}. How many of the three would answer Yes? Answer with
exactly one digit: 0, 1, 2 or 3.

\medskip
\textbf{Arabic}

\begin{otherlanguage}{arabic}
النظام: أنت مُصنِّف خبير لمحتوى وسائل التواصل الاجتماعي باللغة العربية.
أجب برقم واحد فقط.

المستخدم: حكم ثلاثة مصنّفين ناطقين بالعربية، كلٌّ على حدة، على ما إذا
كان التعليق التالي متحيّزًا جنسيًا. يُعدّ التعليق متحيّزًا جنسيًا إذا كان يحطّ من قيمة المرأة، أو يميّز ضدها، أو
يكرّس صورًا نمطية عنها، أو يبرّر الأذى الواقع عليها، سواء كان موجّهًا للنساء
كفئة أو لامرأة بعينها بسبب كونها امرأة. التعليق: [comment]. كم واحدًا منهم سيجيب
بنعم؟ أجب برقم واحد فقط: 0 أو 1 أو 2 أو 3.
\end{otherlanguage}

\subsection*{Hard label, few-shot}

\textbf{English}

\textit{System:} as in the zero-shot hard-label prompt.

\textit{User:} Here are previous judgements on Arabic comments, decided
by majority vote of three annotators. A comment is sexist if it demeans women, discriminates against them,
reinforces stereotypes about them, or justifies harm done to them, whether
directed at women as a group or at an individual woman because she is a
woman. \texttt{\{examples\}}
Comment: \texttt{\{comment\}} Answer:

Each example has the form \textit{Comment: \ldots{} Answer: Yes/No}.

\medskip
\textbf{Arabic}

\begin{otherlanguage}{arabic}
النظام: كما في البرومبت الصفري للتصنيف الثنائي.

المستخدم: هل التعليق العربي متحيّز جنسيًا؟ يُعدّ التعليق متحيّزًا جنسيًا إذا كان يحطّ من قيمة المرأة، أو يميّز ضدها، أو
يكرّس صورًا نمطية عنها، أو يبرّر الأذى الواقع عليها، سواء كان موجّهًا للنساء
كفئة أو لامرأة بعينها بسبب كونها امرأة. وهذه أحكام مصنّفين على تعليقات
سابقة: [examples] التعليق: [comment] الإجابة:
\end{otherlanguage}

\subsection*{Soft label, few-shot}

\textbf{English}

\textit{System:} as in the zero-shot soft-label prompt.

\textit{User:} Three Arabic-speaking annotators independently judged
whether each comment is sexist. A comment is sexist if it demeans women, discriminates against them,
reinforces stereotypes about them, or justifies harm done to them, whether
directed at women as a group or at an individual woman because she is a
woman. Here are their judgements on
previous comments: \texttt{\{examples\}} Comment: \texttt{\{comment\}}
Number who said Yes:

Each example has the form \textit{Comment: \ldots{} Number who said
Yes: 0--3}.

\medskip
\textbf{Arabic}

\begin{otherlanguage}{arabic}
النظام: كما في البرومبت الصفري للتصنيف المرن.

المستخدم: حكم ثلاثة مصنّفين ناطقين بالعربية، كلٌّ على حدة، على ما إذا
كان التعليق متحيّزًا جنسيًا. يُعدّ التعليق متحيّزًا جنسيًا إذا كان يحطّ من قيمة المرأة، أو يميّز ضدها، أو
يكرّس صورًا نمطية عنها، أو يبرّر الأذى الواقع عليها، سواء كان موجّهًا للنساء
كفئة أو لامرأة بعينها بسبب كونها امرأة. وهذه أحكامهم على تعليقات سابقة: [examples]
التعليق: [comment] عدد من أجاب بنعم:
\end{otherlanguage}

\subsection*{Perspectivist, few-shot}

\textbf{English}

\textit{System:} You are simulating the labelling behaviour of one
specific annotator of Arabic social media content. Answer with one word
only.

\textit{User:} Below are the judgements of ONE annotator on Arabic
comments. A comment is sexist if it demeans women, discriminates against them,
reinforces stereotypes about them, or justifies harm done to them, whether
directed at women as a group or at an individual woman because she is a
woman. Annotators differ in where they draw the line, so judge
as THIS annotator would, not as you would. Their previous judgements:
\texttt{\{examples\}} Comment: \texttt{\{comment\}} Answer:

The examples are the annotator's own judgements.

\medskip
\textbf{Arabic}

\begin{otherlanguage}{arabic}
النظام: أنت تحاكي سلوك مصنِّف واحد بعينه لمحتوى وسائل التواصل الاجتماعي
باللغة العربية. أجب بكلمة واحدة فقط.

المستخدم: في ما يلي أحكام مصنِّف واحد على تعليقات عربية. يُعدّ التعليق متحيّزًا جنسيًا إذا كان يحطّ من قيمة المرأة، أو يميّز ضدها، أو
يكرّس صورًا نمطية عنها، أو يبرّر الأذى الواقع عليها، سواء كان موجّهًا للنساء
كفئة أو لامرأة بعينها بسبب كونها امرأة. يختلف المصنّفون في
موضع الخط الذي يرسمونه، فاحكم كما يحكم هذا المصنّف، لا كما تحكم أنت.
أحكامه السابقة: [examples] التعليق: [comment] الإجابة:
\end{otherlanguage}

\section{Confidence intervals for the learning paradigms}
\label{app:ci}

Table~\ref{tab:app-ci-paradigm} gives the difference between each
disagreement-aware encoder and the corresponding hard-label encoder, with
95\% bootstrap confidence intervals, for Macro-F1 and for MD on all, contested
and unanimous comments.

\begin{table}[!ht]
\centering
\scriptsize
\setlength{\tabcolsep}{3pt}
\begin{tabular}{llcccc}
\toprule
 & & \textbf{Macro-F1} & \textbf{MD (all)} & \textbf{MD (contested)} & \textbf{MD (unanimous)} \\
\textbf{Model} & \textbf{Test} & & & & \\
\midrule
\multicolumn{6}{l}{\textit{Soft $-$ hard}} \\
MARBERT & SA & $-$0.010 [$-$0.041, 0.021] & 0.004 [$-$0.027, 0.034] & \textbf{$-$0.169 [$-$0.215, $-$0.127]} & \textbf{0.092 [0.054, 0.129]} \\
 & EG & $-$0.019 [$-$0.056, 0.016] & \textbf{$-$0.030 [$-$0.059, $-$0.001]} & \textbf{0.067 [0.009, 0.124]} & \textbf{$-$0.062 [$-$0.095, $-$0.030]} \\
 & MA & 0.019 [$-$0.014, 0.052] & \textbf{$-$0.048 [$-$0.079, $-$0.017]} & \textbf{$-$0.122 [$-$0.167, $-$0.081]} & $-$0.011 [$-$0.049, 0.030] \\
 & Mix & 0.011 [$-$0.007, 0.029] & $-$0.007 [$-$0.028, 0.014] & \textbf{$-$0.258 [$-$0.286, $-$0.229]} & \textbf{0.104 [0.080, 0.128]} \\
\addlinespace
AraBERT & SA & $-$0.013 [$-$0.039, 0.012] & 0.009 [$-$0.010, 0.028] & \textbf{$-$0.088 [$-$0.123, $-$0.052]} & \textbf{0.058 [0.037, 0.081]} \\
 & EG & $-$0.005 [$-$0.037, 0.024] & \textbf{0.019 [0.0001, 0.039]} & \textbf{$-$0.104 [$-$0.145, $-$0.063]} & \textbf{0.060 [0.040, 0.082]} \\
 & MA & $-$0.015 [$-$0.036, 0.004] & 0.007 [$-$0.013, 0.027] & \textbf{$-$0.133 [$-$0.166, $-$0.100]} & \textbf{0.077 [0.055, 0.098]} \\
 & Mix & $-$0.004 [$-$0.019, 0.012] & $-$0.011 [$-$0.029, 0.006] & \textbf{$-$0.214 [$-$0.237, $-$0.189]} & \textbf{0.078 [0.059, 0.096]} \\
\midrule
\multicolumn{6}{l}{\textit{Persp. $-$ hard}} \\
MARBERT & SA & $-$0.025 [$-$0.057, 0.005] & 0.014 [$-$0.025, 0.053] & \textbf{0.079 [0.013, 0.144]} & $-$0.019 [$-$0.067, 0.030] \\
 & EG & 0.007 [$-$0.029, 0.042] & \textbf{$-$0.054 [$-$0.100, $-$0.010]} & \textbf{0.135 [0.065, 0.211]} & \textbf{$-$0.117 [$-$0.167, $-$0.063]} \\
 & MA & $-$0.004 [$-$0.036, 0.028] & $-$0.029 [$-$0.069, 0.011] & $-$0.018 [$-$0.092, 0.053] & $-$0.035 [$-$0.084, 0.015] \\
 & Mix & 0.008 [$-$0.012, 0.027] & $-$0.024 [$-$0.049, 0.002] & \textbf{$-$0.086 [$-$0.128, $-$0.045]} & 0.004 [$-$0.027, 0.035] \\
\addlinespace
AraBERT & SA & 0.006 [$-$0.026, 0.038] & 0.000 [$-$0.036, 0.036] & \textbf{0.075 [0.009, 0.138]} & $-$0.038 [$-$0.082, 0.006] \\
 & EG & 0.011 [$-$0.024, 0.042] & \textbf{$-$0.059 [$-$0.095, $-$0.021]} & \textbf{0.092 [0.020, 0.158]} & \textbf{$-$0.109 [$-$0.151, $-$0.065]} \\
 & MA & 0.002 [$-$0.026, 0.032] & $-$0.019 [$-$0.056, 0.017] & $-$0.028 [$-$0.092, 0.033] & $-$0.014 [$-$0.059, 0.030] \\
 & Mix & $-$0.012 [$-$0.030, 0.007] & $-$0.012 [$-$0.033, 0.010] & $-$0.023 [$-$0.059, 0.012] & $-$0.006 [$-$0.034, 0.021] \\
\bottomrule
\end{tabular}
\caption{Differences between paradigms with 95\% bootstrap
confidence intervals (2,000 paired resamples, stratified by dialect). Each
cell gives the disagreement-aware model minus the hard-label model; a
negative MD difference means closer alignment with the annotators. All values
are computed on the seed-averaged predictions, so they differ from the
differences between the seed means in Table~\ref{tab:res-main}. Intervals
that exclude zero are in bold.}
\label{tab:app-ci-paradigm}
\end{table}

\section{Effect of prompt language and few-shot examples}
\label{app:prompt-effects}

Figure~\ref{fig:prompt-effects} shows the effect of the prompt language
(English or Arabic) and of few-shot examples on each LLM and test set. For the
hard-label prompt, adding examples increased Macro-F1 in 35 of the 40
combinations of model, language and test set, and the English prompt exceeded
the Arabic one in 34 of the 40. For the soft-label prompt, no single condition
was best: the English prompt gave the lower MD in only 13 of the 40
combinations, six of them for GPT-5.6 Terra, and for the four Arabic LLMs the
Arabic prompt gave the lower MD in most of them. 

Replies outside the requested range of 0--3, which were treated as
unparseable, came from the few-shot soft-label prompt: with the Arabic
prompt, Jais-2 gave such counts for 8--9\% of the Saudi and Moroccan
comments, and Fanar for 35\% of the mixed-dialect comments, in addition to
the 71\% with the English prompt (Section~\ref{sec:res-main}). Fanar's
replies on the mixed-dialect test set were mostly 8, 9 or 10, and those of
Jais-2 were 4. Fanar's out-of-range replies occurred almost only with the mixed-dialect
prompt, which contains 24 examples rather than 8 (two such replies on the
in-dialect test sets with the English prompt, against 958 on the
mixed-dialect test set); we did not identify their cause.

Table~\ref{tab:app-prompt} gives the scores averaged over the three
in-dialect test sets.

\begin{center}
\includegraphics[width=\textwidth,height=0.55\textheight,keepaspectratio]{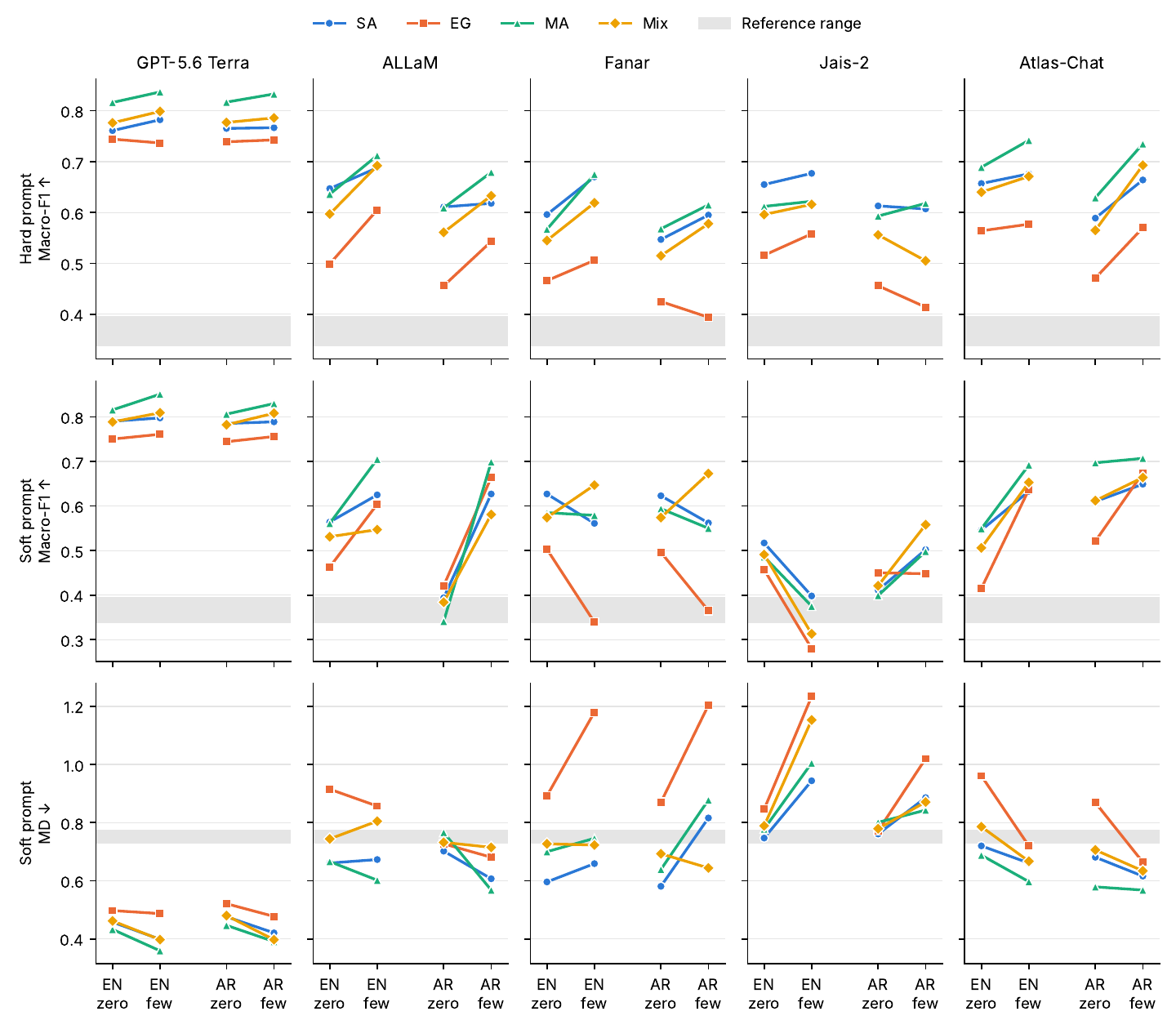}
\captionof{figure}{Effect of the prompt language (English, EN; Arabic, AR) and
of few-shot examples on the LLMs, on each test set. Lines connect the
zero-shot and few-shot results of the same language. The grey band spans the
reference values of the four test sets: the majority class for Macro-F1 and
the constant median of $s$ for MD. GPT-5.6 Terra scores are means over three
runs.}
\label{fig:prompt-effects}
\end{center}

\begin{table}[htbp]
\centering
\small
\setlength{\tabcolsep}{4pt}
\begin{tabular}{llcccccccc}
\toprule
 & & \multicolumn{4}{c}{\textbf{Macro-F1} $\uparrow$} & \multicolumn{4}{c}{\textbf{MD} $\downarrow$} \\
\cmidrule(lr){3-6}\cmidrule(lr){7-10}
 & & \multicolumn{2}{c}{English} & \multicolumn{2}{c}{Arabic} & \multicolumn{2}{c}{English} & \multicolumn{2}{c}{Arabic} \\
\cmidrule(lr){3-4}\cmidrule(lr){5-6}\cmidrule(lr){7-8}\cmidrule(lr){9-10}
\textbf{Model} & \textbf{Paradigm} & Zero & Few & Zero & Few & Zero & Few & Zero & Few \\
\midrule
GPT-5.6 Terra & Hard & 0.774 & 0.785 & 0.774 & 0.781 & 0.511 & 0.493 & 0.507 & 0.504 \\
 & Soft & 0.785 & 0.803 & 0.779 & 0.792 & 0.463 & 0.415 & 0.481 & 0.430 \\
 & Persp. & -- & 0.798 & -- & 0.785 & -- & 0.427 & -- & 0.464 \\
\addlinespace
ALLaM & Hard & 0.594 & 0.668 & 0.559 & 0.613 & 0.820 & 0.693 & 0.872 & 0.801 \\
 & Soft & 0.529 & 0.645 & 0.385 & 0.663 & 0.747 & 0.711 & 0.732 & 0.619 \\
 & Persp. & -- & 0.644 & -- & 0.553 & -- & 0.670 & -- & 0.694 \\
\addlinespace
Fanar & Hard & 0.543 & 0.617 & 0.513 & 0.535 & 0.889 & 0.791 & 0.932 & 0.912 \\
 & Soft & 0.572 & 0.493$^{\dagger}$ & 0.571 & 0.492$^{\dagger}$ & 0.729 & 0.862$^{\dagger}$ & 0.697 & 0.966$^{\dagger}$ \\
 & Persp. & -- & 0.586 & -- & 0.607 & -- & 0.809 & -- & 0.762 \\
\addlinespace
Jais-2 & Hard & 0.594 & 0.619 & 0.554 & 0.546 & 0.826 & 0.776 & 0.878 & 0.904 \\
 & Soft & 0.487 & 0.351$^{\dagger}$ & 0.420 & 0.483$^{\dagger}$ & 0.790 & 1.061$^{\dagger}$ & 0.778 & 0.916$^{\dagger}$ \\
 & Persp. & -- & 0.568 & -- & 0.437 & -- & 0.820 & -- & 0.971 \\
\addlinespace
Atlas-Chat & Hard & 0.637 & 0.665 & 0.563 & 0.656 & 0.749 & 0.710 & 0.866 & 0.726 \\
 & Soft & 0.503 & 0.654 & 0.609 & 0.677 & 0.790 & 0.659 & 0.710 & 0.616 \\
 & Persp. & -- & 0.674 & -- & 0.661 & -- & 0.588 & -- & 0.605 \\
\bottomrule
\end{tabular}
\caption{Effect of prompt language and of few-shot examples on the LLMs.
Scores are averaged over the three in-dialect test sets. The perspectivist
prompts are few-shot only. GPT-5.6 Terra scores are means over three runs.
$^{\dagger}$~Fewer than 100\% valid replies on at least one test set.}
\label{tab:app-prompt}
\end{table}

\section{Sensitivity to the definition of sexism in the prompt}
\label{app:definition}

All prompts in the paper use the definition of sexism given to the annotators
(Section~\ref{sec:ds-annotation}), which is specific to sexism against women.
To examine how sensitive the predictions of an LLM are to the scope of this
definition, we also ran GPT-5.6 Terra with a gender-neutral definition that
does not mention women: \emph{a comment is sexist if it demeans,
stereotypes, objectifies, or discriminates against someone on the basis of
their gender, or endorses such treatment}. 

The comparison used the English
zero-shot and few-shot hard-label prompts on the mixed-dialect test set, with
everything else identical, including the few-shot examples (8 from each
dialect, 24 in total), and three runs per condition
(Table~\ref{tab:app-definition}). With the annotators' definition, false positives were 12.8\% and 13.3\%
lower with the zero-shot and few-shot prompts, while recall changed by about one
percentage point or less. On the majority vote over runs, the difference in
Macro-F1 (0.036 and 0.025) and in false positives (16\% and 17\%) had 95\%
confidence intervals excluding zero (paired bootstrap, 2,000 resamples
stratified by dialect). 

The effect varied across the three dialects: false
positives were 25--26\% lower on the Moroccan comments but not lower on the
Saudi comments. In the qualitative coding (Section~\ref{sec:err-qualitative}),
criticism of men accounted for 22 of the 115 false positives of GPT-5.6 Terra
only with the gender-neutral definition, and for one of the 89 with the
annotators' definition. A broader definition therefore increases the
over-prediction, but substantial over-prediction remains with a definition
aligned to the annotators'.

The comparison was restricted to GPT-5.6 Terra, the strongest-performing LLM
in our evaluation, which obtained the highest Macro-F1 on every test set with
at least one of its prompts (Table~\ref{tab:res-main}); the Arabic-centric
LLMs, which over-predicted the sexist class on most test sets
(Section~\ref{sec:err-direction}), were not included.

\begin{table}[htbp]
\centering
\small
\begin{tabular}{llcccccc}
\toprule
\textbf{Prompt} & \textbf{Definition} & \textbf{Macro-F1} & \textbf{Precision} & \textbf{Recall} & \textbf{FP} & \textbf{FN} & \textbf{Ratio} \\
\midrule
Zero-shot & Gender-neutral     & 0.747 & 0.649 & 0.875 & 269.7 & 71.3 & 1.35 \\
          & Annotators' (main) & 0.776 & 0.682 & 0.885 & 235.3 & 65.3 & 1.30 \\
\midrule
Few-shot  & Gender-neutral     & 0.780 & 0.682 & 0.897 & 238.0 & 58.7 & 1.31 \\
          & Annotators' (main) & 0.799 & 0.710 & 0.888 & 206.3 & 64.0 & 1.25 \\
\bottomrule
\end{tabular}
\caption{GPT-5.6 Terra with the gender-neutral definition of sexism in the
prompt and with the definition given to the annotators, used in the rest of
the paper, on the mixed-dialect test set (English hard-label prompts).
Precision and recall refer to the sexist class. \textit{Ratio}: number of
comments predicted as sexist divided by the number whose majority label is
sexist. The table reports means over three runs; the bootstrap confidence
intervals in the text are based on the majority vote over runs.}
\label{tab:app-definition}
\end{table}

\section{Results without the gold items}
\label{app:nogold}
 
The test sets include 90 gold items, real comments used for quality control
during annotation and chosen because their label follows clearly from the
guidelines. To check that they do not drive the main results, the analyses
were repeated on the 1,257 test comments that are not gold items
(Table~\ref{tab:app-nogold}). As in the error analysis
(Section~\ref{sec:error-analysis}), each model is represented by a single
decision per comment (the mean of the seeds' probabilities for the encoders
and the majority vote over runs for GPT-5.6 Terra), rather than by the mean
Macro-F1 over seeds or runs reported in Table~\ref{tab:res-main}; both columns
of Table~\ref{tab:app-nogold} use the same decisions, so their difference
reflects only the removal of the gold items. Without them, Macro-F1 fell by 0.002--0.021,
consistent with gold items being clear cases, and the contested share rose
from 30.7\% to 32.5\%. The difference in error rate between contested and
unanimous comments changed by at most 0.028, and its 95\% confidence interval
still excluded zero for the hard-label encoders and GPT-5.6 Terra on every
test set, although only narrowly for GPT-5.6 Terra on the Egyptian test set
(lower bound 0.0005). GPT-5.6 Terra still predicted the sexist class more often than the
majority (ratios of 1.12--1.53), and its difference in Macro-F1 from AraBERT on
the mixed-dialect test set remained small (0.022 [$-$0.009, 0.051] on these
decisions, and 0.024 [$-$0.004, 0.051] on the means over runs and seeds);
unlike on the full test set, where the interval on the means over runs and
seeds excluded zero, both intervals included zero.
 
\begin{table}[!ht]
\centering

\begin{tabular}{lcccccc}
\toprule
 & \multicolumn{2}{c}{\textbf{Macro-F1}} & \multicolumn{2}{c}{\textbf{Err.\ con.\ $-$ una.}} & \multicolumn{2}{c}{\textbf{Ratio}} \\
\cmidrule(lr){2-3}\cmidrule(lr){4-5}\cmidrule(lr){6-7}
\textbf{Model} & All & No gold & All & No gold & All & No gold \\
\midrule
MARBERT              & 0.768 & 0.757 & 0.189 & 0.186 & 0.94 & 0.95 \\
AraBERT              & 0.785 & 0.775 & 0.198 & 0.193 & 0.95 & 0.95 \\
GPT-5.6 Terra (few)  & 0.808 & 0.797 & 0.205 & 0.199 & 1.24 & 1.27 \\
GPT-5.6 Terra (zero) & 0.784 & 0.774 & 0.170 & 0.165 & 1.28 & 1.31 \\
ALLaM                & 0.692 & 0.681 & 0.134 & 0.126 & 1.15 & 1.17 \\
Fanar                & 0.619 & 0.604 & 0.070 & 0.054 & 1.67 & 1.71 \\
Jais-2               & 0.616 & 0.603 & 0.096 & 0.081 & 1.23 & 1.24 \\
Atlas-Chat           & 0.671 & 0.656 & 0.112 & 0.101 & 1.28 & 1.30 \\
\bottomrule
\end{tabular}
\caption{Results on the mixed-dialect test set with all 1,347
comments and without the 90 gold items (1,257 comments), for the hard-label
decisions used in the error analysis (Section~\ref{sec:error-analysis}).
\textit{Err.\ con.\ $-$ una.}: error rate on contested minus error rate on
unanimous comments. \textit{Ratio}: comments predicted as sexist divided by
comments whose majority label is sexist. The encoders threshold the mean of
the seeds' probabilities, and GPT-5.6 Terra takes the majority vote over its
three runs, so Macro-F1 differs slightly from Table~\ref{tab:res-main}.}
\label{tab:app-nogold}
\end{table}

\section{Agreement of the models with each annotator}
\label{app:annotators}

Table~\ref{tab:app-annotators} gives the share of test comments on which the
hard-label decision of each model matched the label of each annotator.
A1--A3 are identifiers within each dialect: each dialect was labelled by its
own three annotators, so A1 of the Saudi test set is not the same person as A1
of the Egyptian or Moroccan test set. Within a dialect, each identifier refers
to the same annotator throughout the dataset.

\begin{table}[!ht]
\centering

\small
\setlength{\tabcolsep}{4pt}
\begin{tabular}{lccccccccc}
\toprule
 & \multicolumn{3}{c}{\textbf{Saudi}} & \multicolumn{3}{c}{\textbf{Egyptian}} & \multicolumn{3}{c}{\textbf{Moroccan}} \\
\cmidrule(lr){2-4}\cmidrule(lr){5-7}\cmidrule(lr){8-10}
\textbf{Model} & A1 & A2 & A3 & A1 & A2 & A3 & A1 & A2 & A3 \\
\midrule
\textit{Labelled sexist (\%)} & 46.9 & 46.2 & 36.9 & 34.2 & 35.3 & 30.2 & 34.8 & 53.3 & 54.0 \\
\midrule
MARBERT & 72.6 & 73.7 & 75.4 & 79.1 & 77.9 & 73.3 & 69.9 & 71.6 & 72.7 \\
AraBERT & 74.3 & 75.4 & 72.3 & 76.7 & 77.0 & 71.9 & 75.5 & 76.3 & 75.3 \\
GPT-5.6 Terra & 72.8 & 80.1 & 73.5 & 75.3 & 74.2 & 66.3 & 73.1 & 83.0 & 84.5 \\
ALLaM & 66.8 & 69.2 & 66.6 & 60.0 & 60.7 & 56.5 & 66.7 & 69.7 & 71.6 \\
Fanar & 68.6 & 68.4 & 60.4 & 51.6 & 50.9 & 46.7 & 56.1 & 69.9 & 71.0 \\
Jais-2 & 67.5 & 66.8 & 70.4 & 55.6 & 56.7 & 50.2 & 62.2 & 61.3 & 59.8 \\
Atlas-Chat & 63.9 & 68.1 & 65.0 & 59.3 & 55.8 & 54.4 & 69.7 & 73.1 & 70.8 \\
\bottomrule
\end{tabular}
\caption{Agreement (\%) between the hard-label decision of each model
and the label of each annotator on the in-dialect test sets. The first row
gives the share of test comments that each annotator labelled sexist. Encoder
decisions threshold the mean of the seeds' probabilities; GPT-5.6 Terra takes
the majority vote over its three runs with the English few-shot prompt; the
other LLMs use the English few-shot hard-label prompt.}
\label{tab:app-annotators}
\end{table}

\clearpage
\end{document}